\RequirePackage{fix-cm}
\documentclass[twocolumn]{svjour3}          
\smartqed  

\usepackage{abstract}
\usepackage{appendix}
\usepackage{amssymb,amsmath,amsthm}
\usepackage{mathrsfs}
\usepackage{graphicx}
\usepackage{epstopdf} 
\usepackage{subfigure}
\usepackage{multicol}
\usepackage[numbers, round, sort]{natbib}
\setcitestyle{authoryear,round}
\usepackage{etoolbox}
\usepackage[colorlinks,
            linkcolor=blue,
            anchorcolor=blue,
            citecolor=blue]{hyperref}
\usepackage{newtxmath}
\usepackage{multirow}
\usepackage{marvosym}
\usepackage{bm}
\usepackage{float}
\usepackage[noend]{algpseudocode}
\usepackage{algorithm}  
\usepackage{tabularx}
\usepackage{booktabs}
\usepackage{multicol}
\usepackage{array} 
\usepackage[table]{xcolor}

\usepackage[hang]{footmisc}
\usepackage{geometry}
\makeatletter
\patchcmd{\NAT@citex}
  {\@citea\NAT@hyper@{%
     \NAT@nmfmt{\NAT@nm}%
     \hyper@natlinkbreak{\NAT@aysep\NAT@spacechar}{\@citeb\@extra@b@citeb}%
     \NAT@date}}
  {\@citea\NAT@nmfmt{\NAT@nm}%
   \NAT@aysep\NAT@spacechar\NAT@hyper@{\NAT@date}}{}{}
\patchcmd{\NAT@citex}
  {\@citea\NAT@hyper@{%
     \NAT@nmfmt{\NAT@nm}%
     \hyper@natlinkbreak{\NAT@spacechar\NAT@@open\if*#1*\else#1\NAT@spacechar\fi}%
       {\@citeb\@extra@b@citeb}%
     \NAT@date}}
  {\@citea\NAT@nmfmt{\NAT@nm}%
   \NAT@spacechar\NAT@@open\if*#1*\else#1\NAT@spacechar\fi\NAT@hyper@{\NAT@date}}
  {}{}
  \def\BState{\State\hskip-\ALG@thistlm}

\makeatother
\definecolor{myGreen}{HTML}{33FF00}
\definecolor{myRed}{HTML}{FF3030}
\definecolor{myGrey}{HTML}{AA5555}
\definecolor{myWhite}{HTML}{FFFFFF}
\definecolor{maroon}{cmyk}{0,0.87,0.68,0.32}
\definecolor{petr}{HTML}{5555FF}
\definecolor{josef}{HTML}{FF3030}

 \journalname{Preprint}
\begin{document}
\begin{sloppypar}
\title{On Procedural Adversarial Noise Attack And Defense
}


\author{Jun Yan,
        Xiaoyang Deng,
         Huilin Yin,
        Wancheng Ge
}
\date{Received: date / Accepted: date}

\institute{Jun Yan \at
              Department of Information and Communication Engineering at Tongji University, Shanghai, China \\
              \email{yanjun@tongji.edu.cn}           
           \and
           Xiaoyang Deng \at
              Department of Control Science and Engineering at Tongji University, Shanghai, China \\
              \email{dxytju@tongji.edu.cn}
            \and
             \Letter Huilin Yin (Corresponding author) \at
             Department of Control Science and Engineering at Tongji University, Shanghai, China \\
             \email{yinhuilin@tongji.edu.cn}
            \and
            Wancheng Ge \at
            Department of Information and Communication Engineering at Tongji University, Shanghai, China \\
            \email{gwc828@tongji.edu.cn}
}


\maketitle
\begin{abstract}
Deep Neural Networks (DNNs) are vulnerable to adversarial examples which would inveigle neural networks to make prediction errors with small perturbations on the input images. However, most of the proposed attacks depend on specific models and data. Researchers have been devoted to promoting the study on universal adversarial perturbations (UAPs) which have little prior knowledge on data distributions. Procedural adversarial noise attack is a data-independent universal perturbation generation method. Adding the texture pattern with the shading based on the rendering technology to the original image achieves the deceit on the neural networks without changing the visual semantic representations. Similar to the disturbance of shading on human eyes, the shading generated by procedural noise can also fool the neural networks. Previous research on procedural adversarial noise provides a baseline, however, the performance of attack can be promoted with a more aesthetic rendering effect. In this paper, we propose two universal adversarial perturbation (UAP) generation methods based on procedural noise functions: Simplex noise and Worley noise. The UAPs with such solid textures realize the cross-model and cross-data attack effect. We provide a detailed empirical study to validate the effectiveness of our method. In the attack experiments, the results of our proposed methods surpass the state-of-the-art methods of procedural adversarial noise attack, black-box attack, and universal adversarial perturbation attack on the ImageNet dataset and the CIFAR-10 dataset. Moreover, before our work, there is no discussion about the defense on the procedural adversarial noise attacks. Therefore, we evaluate the denoising-based defense methods and other state-of-the-art defense methods on ImageNet and CIFAR-10. The result of the defense experiment verifies some theoretical analysis of robustness in deep learning. Code is available at \url{https://github.com/momo1986/adversarial_example_simplex_worley}. 
\end{abstract}

\section{Introduction}
\label{intro}
The last decade is an era of deep learning's renaissance.
In the field of computer vision, Convolutional Neural Networks (CNNs)~\citep{AlexNet, NIN, VGG, InceptionV1, InceptionV2, InceptionV3, InceptionV4, ResNet, SENet} have been widely used in real applications related to visual perception and cognition. Using automated machine learning (AutoML) to replace craft neural network design is also a popular trend. Zoph and Le~\citep{NAS} proposed a neural architecture search method based on reinforcement learning which outperforms most of the CNN models on the metrics of the prediction accuracy.

However, deep learning cannot guarantee security. Despite the high accuracy of the clean testing dataset, most CNN models are vulnerable to adversarial examples. White-box attacks~\citep{FGSM, BIM, PGD, JSMA, DeepFool, C&W, EOT, BPDA} are gradient-based  to update adversarial perturbations with the exploration of the model structures during the optimization process. Other black-box attacks~\citep{NATTACK, NES, SPSA, BoundaryAttack, SimpleAttack, ParsimoniousAttack} are built on lots of queries of input information and output information of models, which is time-consuming. To make the attack convenient in the deployment, researchers are devoted to finding image-agnostic adversarial perturbations. Universal adversarial perturbations (UAP) introduced in the previous work~\citep{UAP} can fool state-of-the-art image classification models with high possibilities and small perturbations. The proposed UAP methods are quasi-imperceptible and do not require solving an optimization problem. The universal perturbations can transfer between different images and different models. Afterward, a lot of papers~\citep{UAN, NAG, FastFeatureFool, GD_UAP} are published. 

Generating universal adversarial examples based on the procedural noise functions can be a research direction. These procedural noise functions are commonly used in computer graphics and designed to be parametrizable, customizable, and aesthetic~\citep{OldGabor2}. Adding textures and patterns in the pictures does not modify the visual semantic representations. The perturbation patterns generated by procedural noise functions have similar structures with the existing universal adversarial perturbations~\citep{UAP, SVD_UAP}. Generally speaking, the human perception system would be disturbed by the shadings. The sensibility is similar on the neural networks that the deep visual classifiers would be fooled by the procedural adversarial noises with the shadings. Therefore, adversarial learning on such procedural noises can improve the visual classifier's robustness under the untergeted attack. The perception systems of autonomous vehicles need a performance guarantee when faced with the abnormal scenarios of sensors. The visual content audit system of Internet enterprises should inference correctly in the case of malicious image tampering. Therefore, robustness under the procedural adversarial noise is an explorable research direction. A viewpoint was put forward~\citep{Evaluate_Robustness} that defending random perturbations based on Gaussian noise is a basic requirement. Rayleigh noise, Gamma noise, and Salt-And-Pepper noise are also commonly used noise function models. 
In previous work~\citep{Gabor, Perlin}, two adversarial attacks based on procedural noise functions are proposed with the state-of-the-art effect. 
Nevertheless, many proposed noise attack methods do not have a superior performance which needs further improvement. Currently, Perlin noise attack~\citep{Perlin} is a state-of-the-art procedural adversarial noise attack. However, Perlin noise has several shortcomings: visually significant anisotropy, gradient artifacts, and higher computation cost. The drawbacks of the rendering technology existed in Perlin noise may hinder the adversarial attack performance in computer vision. Therefore, it gives us an inspiration to promote the research on the procedural adversarial noise attack. Moreover, before our work, there is almost no discussion on the defense technologies under the procedural adversarial noise attacks.

In this paper, we propose two universal adversarial perturbation attack methods based on noise functions: Simplex noise attack and Worley noise attack. We empirically demonstrate that the neural networks are fragile to the procedural noises that act as the universal adversarial perturbations (UAPs).
In the attack experiment, our methods show superior performance compared with the state-of-the-art noise attack methods, black-box attack methods, and UAP methods.
In the defense experiment, we evaluate the denoising methods and the defense methods provided by the RealSafe~\citep{RealSafe} benchmark.

Our contributions in this paper are listed as follows:
\begin{itemize}
\item We propose two procedural adversarial noise perturbation attack methods: Simplex noise perturbations and Worley noise perturbations. Such $\ell_{\infty}$-norm attacks surpass state-of-the-art invasion effect on the ImageNet dataset~\citep{ImageNet} and CIFAR-10 dataset~\citep{CIFAR-10}. 
    
\item An empirical and comparative study with other transfer-based black-box attack methods, query-based black-box attack methods, and other universal adversarial perturbation (UAP) methods is made to certify the cross-model attack performance of our procedural adversarial noises.

\item To our best knowledge, we are one of the earliest groups to discuss the defense on the procedural adversarial noise attacks and analyze the associated robustness with the evaluation benchmark.
    
\end{itemize}

This paper is organized as follows. The related works are introduced in Section II. In Section III, our proposed approach is illustrated. Metrics and experiment results comparison are presented in Section VI. Finally, the conclusion is presented in Section V.
\section{Related Work}
\subsection{Black-Box Adversarial Attack, Universal Adversarial Perturbations,  and Procedural Adversarial Noise}
Compared with the white-box adversarial attacks which need prior knowledge of model structures and data distributions, researchers are devoted to the proposal of black-box attack methods. Some black-box adversarial attacks are achieved via the transfer of the white-box attacks~\citep{PGD, BIM, MIM}.  However, the cross-architecture performance cannot be guaranteed. Other black-box adversarial attack methods depending on the query of input/output (I/O) are score-based~\citep{NATTACK, NES, SPSA, SimpleAttack} or decision-based~\citep{BoundaryAttack, ParsimoniousAttack}. Nevertheless, they have large time complexities for query while there is still no guarantee for the cross-structure performance.

The universal adversarial perturbations (UAPs) proposed by Dezfooli et al.~\citep{UAP} are quasi-imperceptible to the human eyes so that the deep neural network can be deceived. Normally, UAPs~\citep{UAP} have geometric correlations between different parts of the decision boundary of the classifier. The vanilla UAP methods and the universal perturbation generation methods based on generative models~\citep{UAN, NAG} are data-driven which limits their further usages. The proposal for data-independent adversarial perturbations is a research focal point. Mopuri et al.~\citep{FastFeatureFool} proposed a data-independent approach to compute universal adversarial perturbations with an efficient and generic objective to construct image-agnostic perturbations to fool CNNs. They also found that misfiring the features in the hidden layers can lead to eventual misclassifications. 
Mopuri et al.~\citep{GD_UAP} proposed a data-independent perturbation generation method that exploits minimal prior information about the training data distribution and extended such technologies to the task of object detection and semantic segmentation. 

Using procedural adversarial noise as data-independent perturbations can be a research direction. The procedural adversarial noise attacks proposed in previous work~\citep{Gabor, Perlin} are inspired by the theoretical research of UAP~\citep{SVD_UAP}. Gabor noise is the convolution of a sparse white noise and a Gabor kernel, making it a type of Sparse Convolution Noise~\citep{OldGabor1, OldGabor2}. Perlin adversarial noise attack~\citep{Perlin} is proposed to generate universal adversarial perturbations based on the lattice gradient noise invented in the computer graphics researches~\citep{OldPerlin1, OldPerlin2}. However, there exist drawbacks in the rendering technologies of Perlin noise which may hinder the adversarial attack performance in computer vision. Therefore, it is necessary to promote further research on procedural adversarial noises. Olano et al.~\citep{Simplex} proposed the Simplex noise function while Worley~\citep{Worley} proposed the Worley noise to realize the graphics rendering function. The pioneering researches in the field of computer graphics inspire us to promote exploration in the field of universal adversarial perturbation related to the study of pattern recognition.

\subsection{Defense}
There are diverse views about the robustness of deep learning. Some researchers are pessimistic that the problems of adversarial examples are inevitable for the distributions with complex image classes in high-dimensional spaces~\citep{AdvInputDimension, Adversarial_Inevitable, AdvSpheres}. Therefore, there is little point in defense under the adversarial perturbations. Dezfooli et al.~\citep{GeoUAP} showed that the flatness property of the neural network's decision boundary can lead to the existence of small perturbations. This work is a theoretical basis of universal adversarial perturbations.

The optimistic view is held in other researches. Besides some theoretical analyses~\citep{AdvVulClf, AdvRisk}, many defense methods are also proposed to improve the robustness and evaluated in a benchmark~\citep{RealSafe}. The state-of-the-art defense methods are adversarial training~\citep{PGD, TRADES, EnsembleAdversarialTraining} whose ``gradient penalty" mechanism boosts the performance of robustness of neural networks under the adversarial attacks. Ross et al.~\citep{advGradientPenalty} analyzed ``gradient penalty" phenomenon from a theoretical perspective. 

In our point of view, the research about the Frequency Principle (F-Principle) of deep learning gives the interpretations on the robustness of neural networks. Xu et al.~\citep{FrequencyXu} held the opinion that the neural networks are inclined to fit the low-frequency elements which is corresponding with the generalization ability of the models. Rahaman et al.~\citep{FrequencyBengio} analyzed the Rectified Linear Unit (ReLU) activation function's dense and smooth property with Stokes Theorem in the topology and concluded that the spectral attenuation of the ReLU function has a strong anisotropy in the high-dimensional space while the upper bound of the ReLU function's Fourier transform amplitude are within the Lipschitz constraint. A different viewpoint was proposed by Weinan E et al.~\citep{FrequencyWeinanE} that high-frequency elements are also important with the mathematical analysis. Making neural networks robust means not giving up the high-frequency elements immediately. Similar conclusions~\citep{FrequencyGoogle, FrequencyCMU} were reached that adversarial training is related to some high-frequency elements and generalization ability is related to the low-frequency elements. 

In this paper, we would make an empirical study on the defense technologies related to F-Principle and defense technologies provided in the released RealSafe~\citep{RealSafe} benchmark to evaluate the robustness under the procedural adversarial noise attacks.
\section{Approach}
In this section, we propose our procedural adversarial noise attack methods.

\subsection{Data-independent Perturbations For Fooling}
\begin{figure}[htbp]
\centering
\includegraphics[scale=0.25]{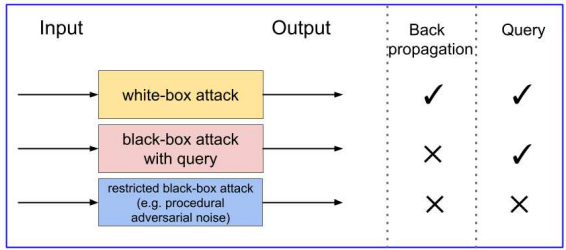}
\caption{Difference between white-box attack, black-box attack with query, and black-box attack without query. Our proposed procedural adversarial noise attack method requires no query of model input/out information.}
\label{fig:blackBoxScenario}
\end{figure}
The basic problem our paper discusses is mainly defined under the theoretical frameworks of UAPs~\citep{UAP}.
It aims to craft an image-agnostic perturbation $\delta \in \mathbb{R}^{d}$ with the procedural noise function to fool the classification of the CNN $f$ on data distribution $X$.
The attack should be satisfied with Eq. (\ref{perturbation}) when attacking the sample image $x \in X$:
\begin{equation}\label{perturbation}
f(x+\delta) \neq f(x), \text { for } x \in X
\end{equation}
The pixel intensities of $\delta \in \mathbb{R}^{d}$ are constrained, noise attack can be regarded as $l_\infty$-norm attack.
\begin{equation}\label{l_infinity_attack}
\begin{array}{c}
f(x+\delta) \neq f(x), \text { for } x \in X \\
\|\delta\|_{\infty}<\xi
\end{array}
\end{equation}
In our paper, the attack defined in the form of Eq. (\ref{l_infinity_attack}) is black-box and data-independent. As illustrated in Fig.~\ref{fig:blackBoxScenario},  our proposed method is gradient-free and requires no prior knowledge of the model structures and data distributions. In contrast, the white-box attack methods are gradient-based while the popular, non-restricted black-box attack methods have the access to input and output information.

\subsection{Simplex Noise Attack}
\begin{figure*}[htbp]
\centering
\subfigure[]{
\centering
\includegraphics[width=0.4\hsize]{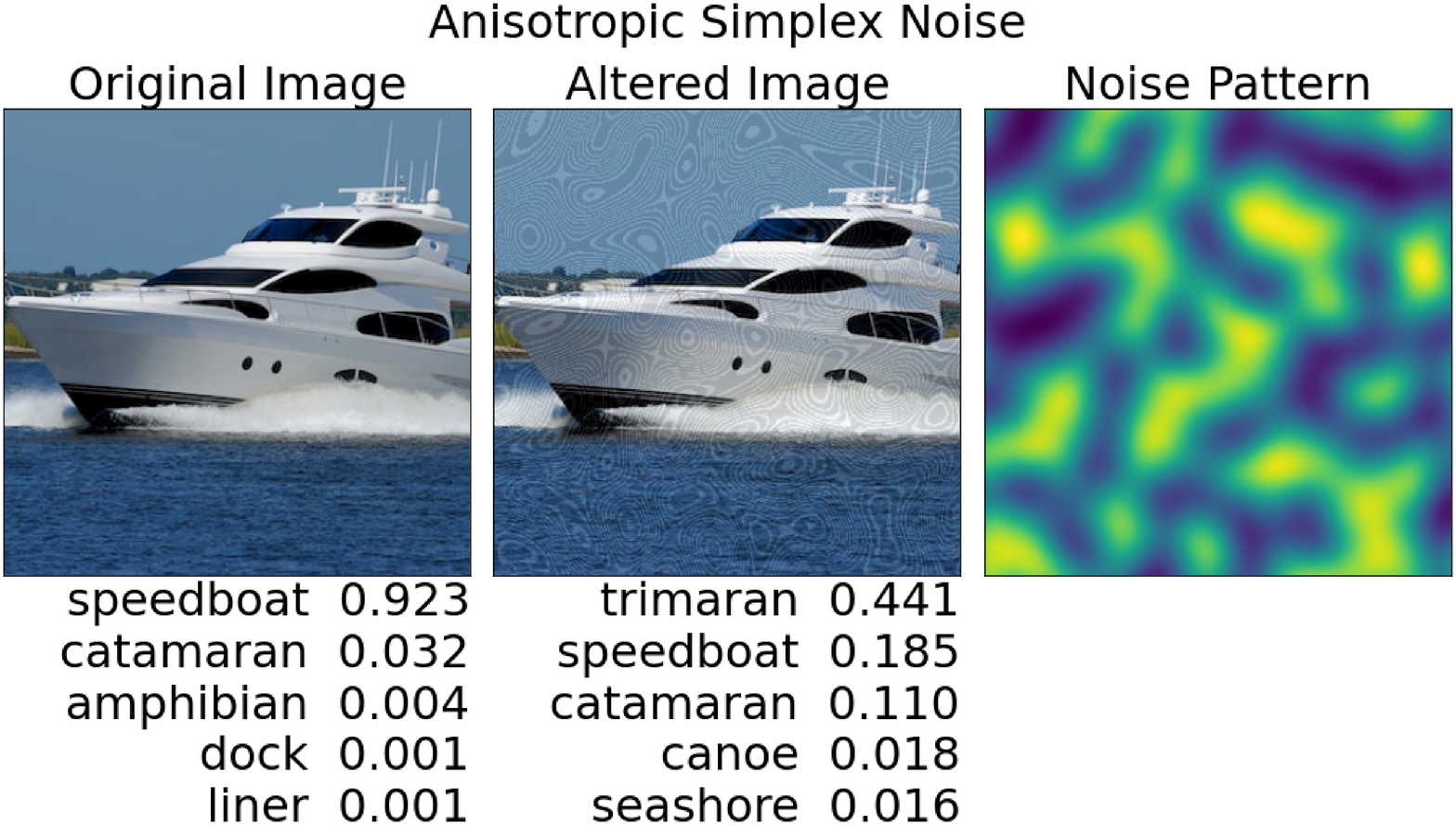}
}%
\subfigure[]{
\centering
\includegraphics[width=0.4\hsize]{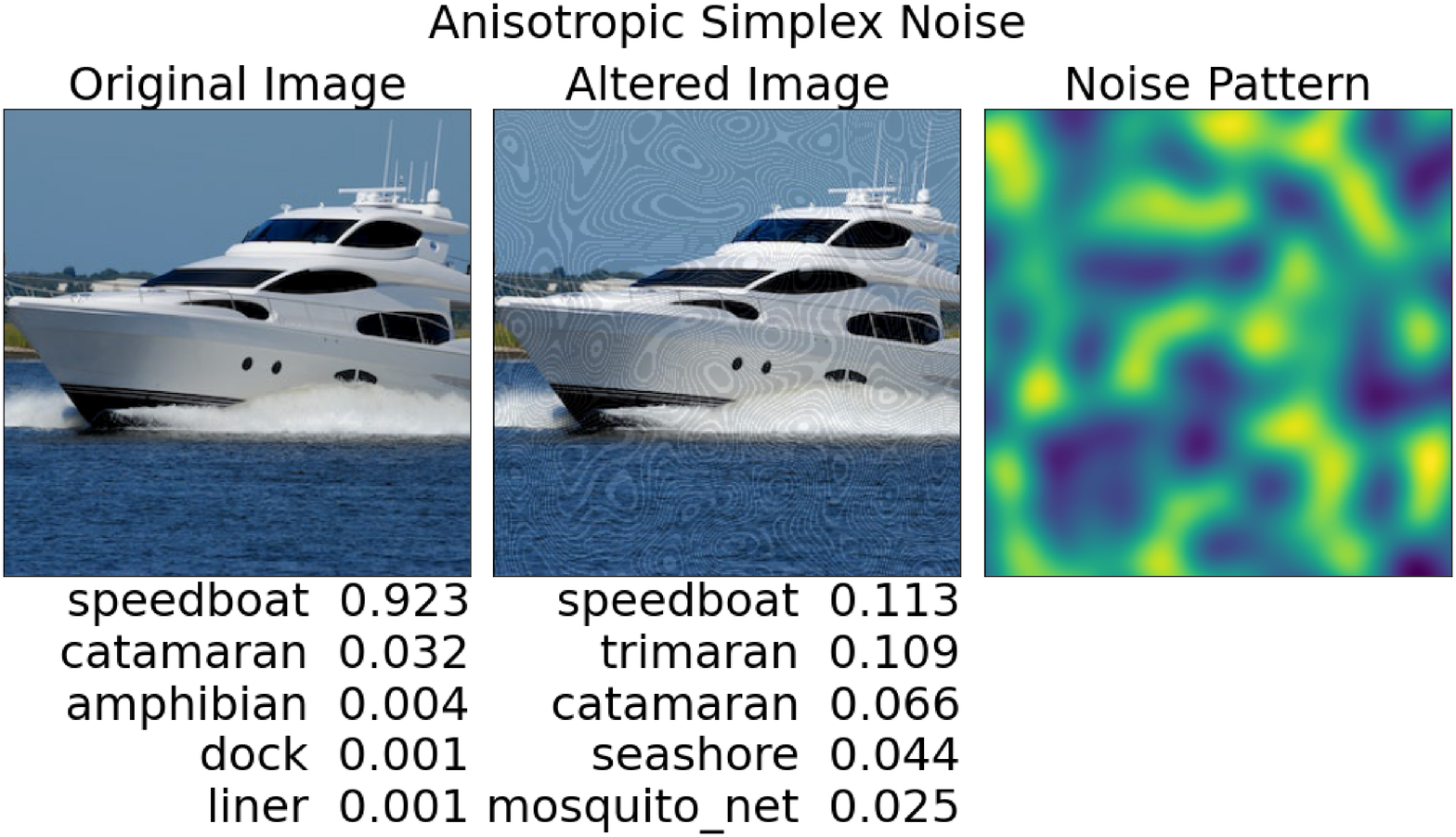}
}%
\quad 
\subfigure[]{
\centering
\includegraphics[width=0.4\hsize]{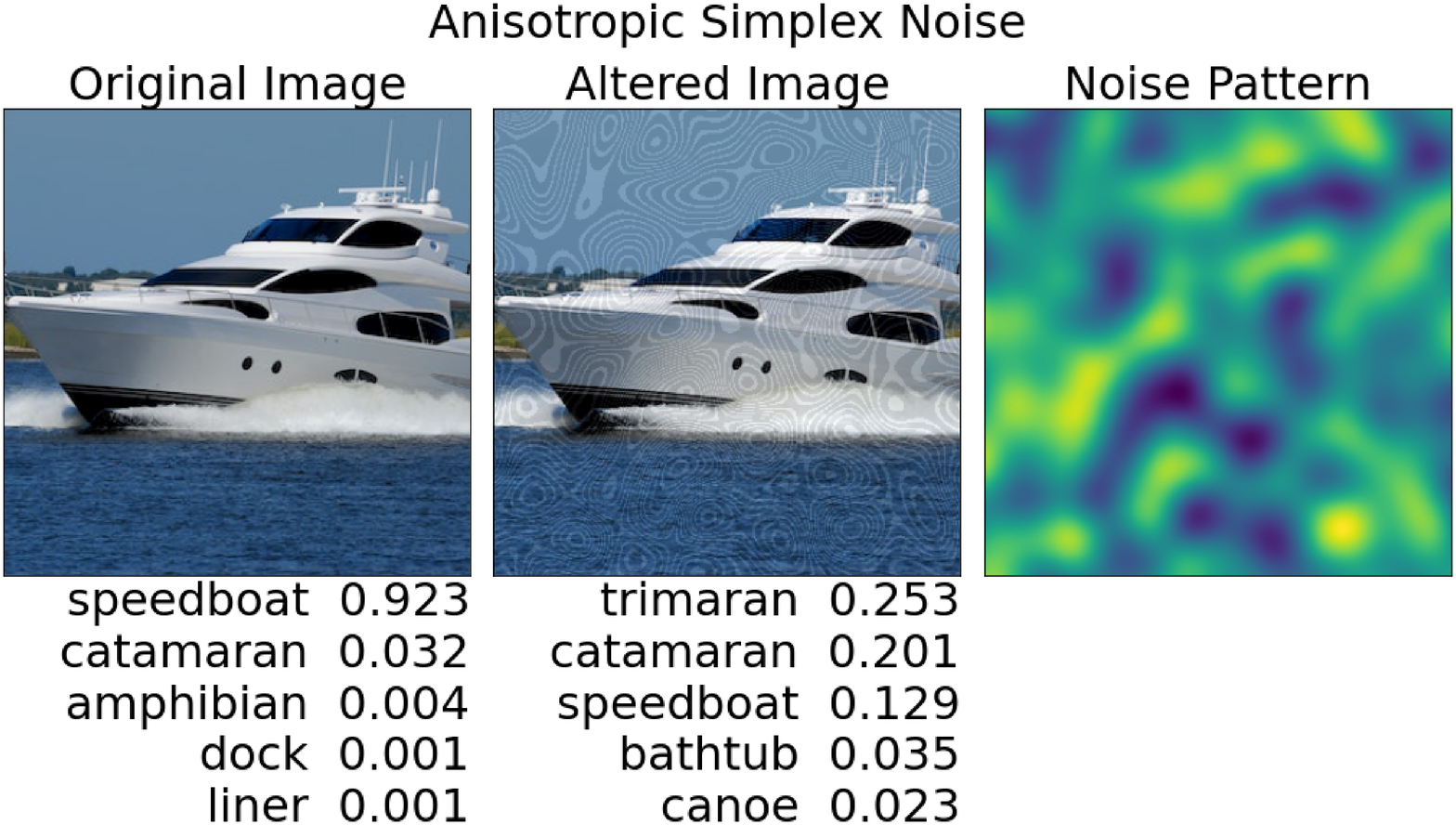}
}%
\caption{Demo of procedural adversarial attack in Simplex noise function on ImageNet dataset.
Fig. (a) illustrates a Simplex noise attack in 2D dimensions, perturbations generated at the iteration step of every 40 pixels with perturbation budget $\varepsilon=0.0465$.
Fig. (b) illustrates a Simplex noise attack in 3D dimensions, perturbations generated at the iteration step of every 40 pixels with perturbation budget $\varepsilon=0.0465$.
Fig. (c) illustrates a Simplex noise attack in 4D dimensions, perturbations generated at the iteration step of every 40 pixels with perturbation budget $\varepsilon=0.0465$.}
\label{fig:SimplexNoise}
\end{figure*}
Simplex noise~\citep{Simplex} can be seen as a variant of Perlin noise whose procedural shading can be better applied to the real-time hardware platforms. Firstly, it has lower computation complexity with fewer multiplications compared to Perlin noise and can be better adapted to the higher dimension. 
Secondly, it has a well-defined and continuous gradient (for almost everywhere) that can be computed quite cheaply. Last but not least, Simplex noise has no noticeable directional artifacts (is visually isotropic) compared to Perlin noise.

The Simplex noise generation procedure would do coordinate skewing according to Eq. (\ref{coordinateSkewing}) to realize input coordinate transform, where $n$ is the dimension number, $F$ is the intermediate variable of the operation.
The vertex arrangement of a hypercubic honeycomb should be squashed along its main diagonal until the distance between the points $(0, 0, ..., 0)$ and $(1, 1, ..., 1)$ is equal to the distance between the points $(0, 0, ..., 0)$ and $(1, 0, ..., 0)$. The variables $u$, $u^{\prime}$ denote the horizontal coordinate while the variable $v$, $v^{\prime}$ denote the vertical coordinate. The $(u,\ v)$ is the coordinate before skew while the $(u^{\prime},\ v^{\prime})$ is the coordinate after skew.
\begin{equation}\label{coordinateSkewing}
\begin{array}{c}
F=\frac{\sqrt{n+1}-1}{n} \\
u^{\prime}=u+(u+v+\cdots) * F \\
v^{\prime}=v+(u+v+\cdots) * F 
\end{array}
\end{equation}
Compared to original Perlin noise using a cubic interpolation grid, Simplex noise uses a grid based on the simplicial subdivision. Then, the simplex vertex is added back to the skewed hypercube's base coordinate and hashed into a pseudo-random gradient direction. It differs from different dimensions~\citep{SimplexReport}. For 2D, 8 or 16 gradients distributed around the unit circle is a good choice. For 3D, the recommended set of gradients is the midpoints of each of the 12 edges of a cube centered on the origin. For 4D, the set of the gradients is formed from the midpoints of each of the 32 edges in a 4D hypercube. After the operation of gradient selection (step 6 of Algorithm \ref{alg:Simplex}), the simplex noise function does kernel summation to get the restored coordinate of each of the vertices according to Eq. (\ref{kernel_summation}), where $n$ is the dimension number, $G$ is the intermediate variable of the operation. The Eq. (\ref{kernel_summation}) realizes the calculation of the position without skew in the normal simplicial coordinate system. The $(u,\ v)$ is the coordinate without skew while the $(u^{\prime},\ v^{\prime})$ is the coordinate with skew:
\begin{equation}\label{kernel_summation}
\begin{array}{c}
G=\frac{1-1 / \sqrt{n+1}}{n} \\
u=u^{\prime}-\left(u^{\prime}+v^{\prime}+\ldots\right) * G \\
v=v^{\prime}-\left(u^{\prime}+v^{\prime}+\ldots\right) * G 
\end{array}
\end{equation}
This unskewed displacement vector is used to compute the extrapolated gradient value using a dot product and calculate the squared distance to the point $d^{2}$. Eq. (\ref{kernel_summation2}) determines each vertex's summed kernel contribution where the variable $r^{2}$ is usually set to either 0.5 or 0.6 in previous work~\citep{Simplex} and the gradient information has been obtained.
\begin{equation}\label{kernel_summation2}
\left(\max \left(0, r^{2}-d^{2}\right)\right)^{4} \cdot(\langle\Delta u, \Delta v, \ldots\rangle \cdot\langle\operatorname{grad} u, \operatorname{grad} v, \ldots\rangle)
\end{equation}
The Simplex noise algorithm is described in Algorithm \ref{alg:Simplex}. The adversarial perturbations generated by  Algorithm \ref{alg:Simplex} is universal which do not depend on models and images.  An implementation typically involves four steps: coordinate skewing, simplicial subdivision, gradient selection, and kernel summation. The images with the Simplex-noise perturbations can fool the neural networks.
\begin{algorithm}[!htbp]
\caption{SIMPLEX$(H, W, S)$}
\label{alg:Simplex}
\hspace*{13mm} {\bf Input:} The image height $H$, image width $W$, iteration step $S$\\
\hspace*{13mm} {\bf Output:}The noise perturbation matrix $\boldsymbol{P}$\\
\begin{algorithmic}[1]
\State Initialize noise matrix $\boldsymbol{P}$
\For{$j=1$ to $H$}
	\For{$i=1$ to $W$} 
	 	\State Do the operation of skewing according to Eq. (\ref{coordinateSkewing}) on the coordinate of $(i/S, j/S)$;
		\State Sort the values of internal coordinates in decreasing order to determine which skewed orthoscheme simplex the point lies in;
		\State Add back to the skewed hypercube's base coordinate to hash into a pseudo-random gradient direction;
		\State Do the operation of unskewing according to Eq. (\ref{kernel_summation});
		\State Get the kernel summation value $k$ based on Eq. (\ref{kernel_summation2});
		\State$P[i, j] = k$
	\EndFor
\EndFor
\State \Return$\boldsymbol{P}$
\end{algorithmic}
\end{algorithm}

Fig. \ref{fig:SimplexNoise} illustrates some qualitative results of the adversarial attack based on the Simplex noise function. The perturbation budget $\varepsilon$ of $\ell_{\infty}$-norm attack is set to $0.0465$ at the iteration step of 40 to generate 2D, 3D, and 4D Simplex noise. As we can see from Fig. \ref{fig:SimplexNoise}, the adversarial attack based on the perturbations generated by Simplex noise can fool the prediction of neural network or at least realize the effect of disturbance on prediction.

\subsection{Worley Noise Attack}
\begin{figure*}[htbp]
\centering
\subfigure[]{
\centering
\includegraphics[width=0.4\hsize]{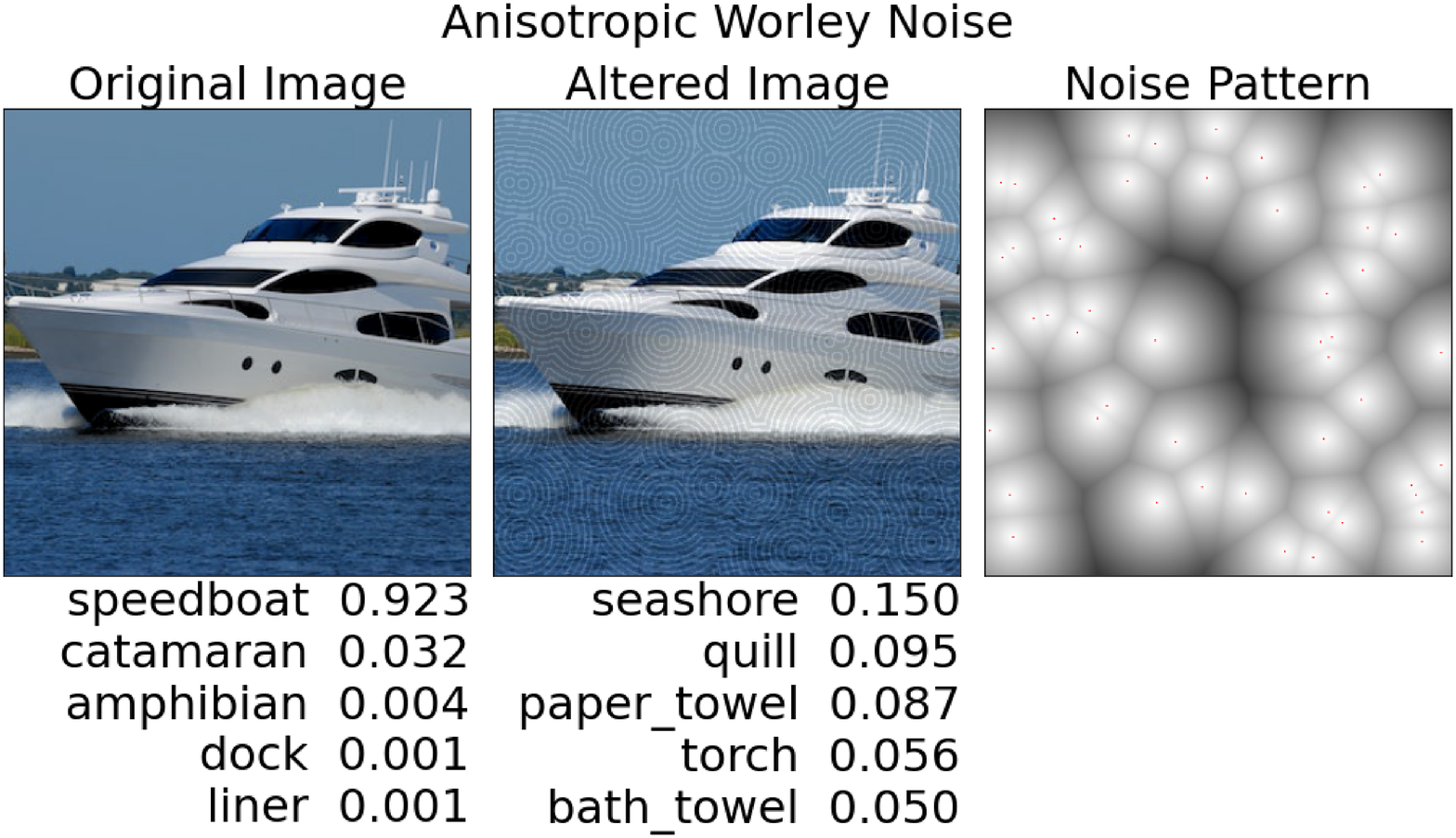}
}%
\subfigure[]{
\centering
\includegraphics[width=0.4\hsize]{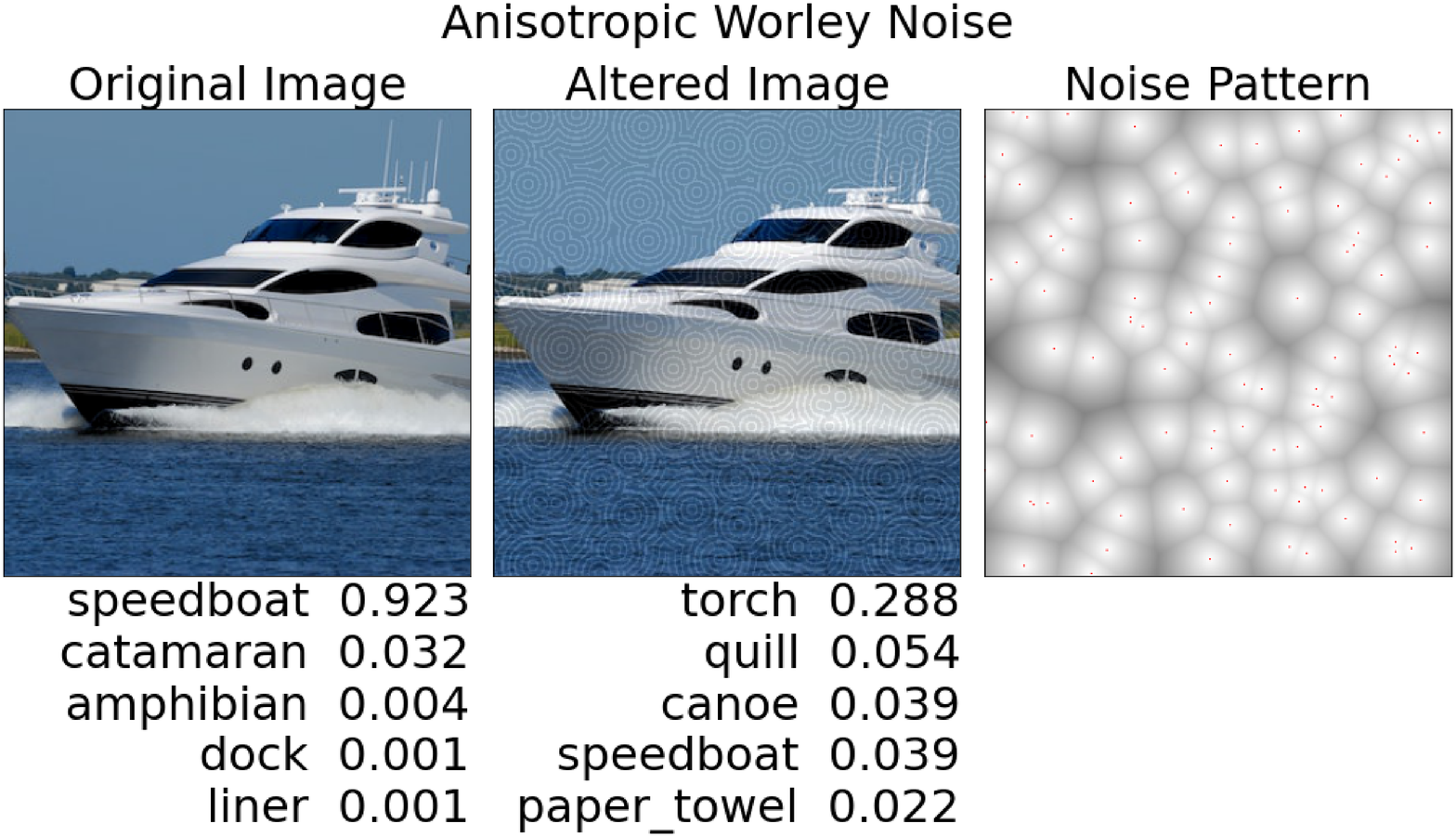}
}%
\caption{Demo of procedural adversarial attack in Worley noise function on ImageNet dataset.
Fig. (a) illustrates Worley noise attack with 50 points perturbed on the $\varepsilon=0.0465$. 
Fig. (b) illustrates Worley noise attack with 100 points perturbed on the $\varepsilon=0.0465$.}
\label{fig:WorleyNoise}
\end{figure*}
Worley noise~\citep{Worley} is generated on the cellular texture in which certain points in 3D space are randomly selected. According to the ``nearest neighbor" principles, returned functions are mapped to colors or texture coordinates.
In the field of computer graphics, this noise function provides solid texturing to the rendered objects. 

Worley noise function defines 3D space partitioned into cubes with faces at integers, in the RGB image scenario, $z=0$. A point $p$ of an index $i$ with the real coordinate $(x_{i},\ y_{i},\ z_{i})$ is selected to generate several feature points inside the cube.
Steps will be repeated until required perturbed point numbers $N$ have been iterated and added into the set. 
According to Eucidean distance defined in Eq. (\ref{EucideanDistance}), cube feature points are selected, calculated, sorted, and checked on the principle of "nearest neighbors".
\begin{equation}\label{EucideanDistance}
d=\sqrt{\sum_{i=1}^{n}\left(x_{i}-y_{i}\right)^{2}}
\end{equation}
Implementation is elucidate by such procedure defined in Algorithm \ref{alg:Worley}. The adversarial perturbations generated by  Algorithm \ref{alg:Worley} is universal which do not depend on models and images.
\begin{algorithm}[!htbp]
\caption{WORLEY$(H, W, N)$}
\label{alg:Worley}
\hspace*{13mm} {\bf Input:} The image height $H$, image width $W$, perturbed point numbers $N$\\
\hspace*{13mm} {\bf Output:} The noise perturbation matrix $\boldsymbol{P}$\\
\begin{algorithmic}[1]
\State Initialize the image grid matrix $\boldsymbol{G}$ according to image width $W$ and image height $H$
\State Select $N$ points of different coordinates randomly in the images with image height $H$, image width $W$ with the set $\boldsymbol{R}$
\For{$j=1$ to $H$}
	\For{$i=1$ to $W$} 
	 	\State Map cube feature points via getting normalized distance from "nearest neighbors" according to Eucidean distance in Eq. (\ref{EucideanDistance}) and get value $m$
		\State $G[j, i] = (m, m, m, 255)$
	\EndFor
\EndFor
\For{$i, j \in R$}
	\State $G[j, i] = (255, 0, 0, 255) $
\EndFor
\State Split $\boldsymbol{P}$ with 4 channels and concatenating with 3 channels in RGB format to generate the matrix $\boldsymbol{P}$
\State \Return$\boldsymbol{P}$
\end{algorithmic}
\end{algorithm}

Fig. \ref{fig:WorleyNoise} illustrates some qualitative results of the adversarial attack based on the Worley noise function. The perturbation budget $\varepsilon$ of $\ell_{\infty}$-norm attack is set to $0.0465$, 50 or 100 points will be randomly clustered.
As we can see from Fig. \ref{fig:WorleyNoise}, the ground truth label is ``speedboat" while the prediction label on the adversarial attack is ``seashore" (50 perturbed points) or ``torch"  (100 perturbed points. Its performance on fooling neural networks is not worse or even better than Simplex noise and other procedural adversarial noises.

\section{Experiment}
In this section, the experiments on procedural adversarial noise attack and defense would be illustrated. 
Our attack experiment and defense experiment with the denoising methods is implemented under the Keras framework, while the defense methods described in RealSafe~\citep{RealSafe} have their corresponding pre-trained models under the framework of vanilla Tensorflow and PyTorch. 
On ImageNet~\citep{ImageNet}, due to the computation limit, we use pre-trained models and only test them on the validation dataset with 50,000 samples. On CIFAR-10~\citep{CIFAR-10}, there are 50,000 training images and 10,000 test images, we implement the training procedure and test on the dataset in the experiment of adversarial attack and denoising defense. 

\subsection{Metrics}
\textbf{Evasion rate} of a perturbation over the dataset can measure the perturbations and we select it as the evaluation metrics for attack. Given model output $f$ , input $x\in X$ with perturbations $s$, and small $\varepsilon>0$, the universal evasion of $s$ over $X$ is defined in Eq. (\ref{attackRate}):
\begin{equation}\label{attackRate}
\frac{|\{x \in X: \arg \max f(x+s) \neq \tau(x)\}|}{|X|}, \quad\|s\|_{\infty} \leq \varepsilon
\end{equation}
where $\tau(x)$ is the true class label of $x$. An $\ell_{\infty}$-norm constraint on $s$ ensures that the perturbations are small and do not drastically change the semantic understanding and representation of the images. When ``evasion rate" is used as a metric for UAP,  it can also be called ``universal evasion rate". This metric is a derivation of previous work~\citep{Perlin}.

In the defense scenario, we just redefine the \textbf{robust accuracy} in Eq. (\ref{robustAccuracy}):
\begin{equation}\label{robustAccuracy}
\frac{|\{x \in X: \arg \max f(x+s) = \tau(x)\}|}{|X|}, \quad\|s\|_{\infty} \leq \varepsilon
\end{equation}
\subsection{Comparison Experiment of Adversarial Noise Attack}
\begin{figure*}[htbp]
\centering
\subfigure[]{
\centering
\includegraphics[width=0.4\hsize]{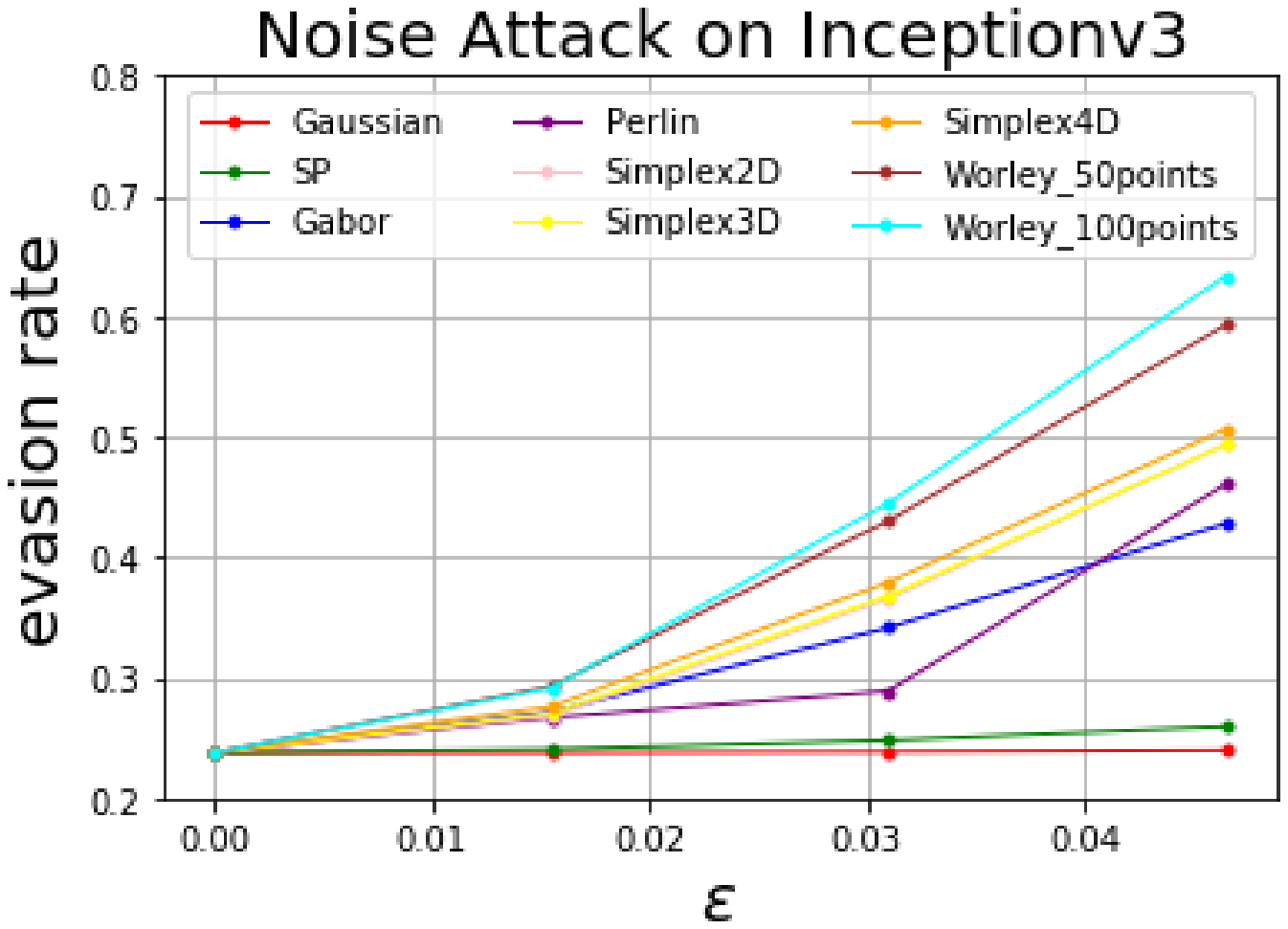}
}%
\subfigure[]{
\centering
\includegraphics[width=0.4\hsize]{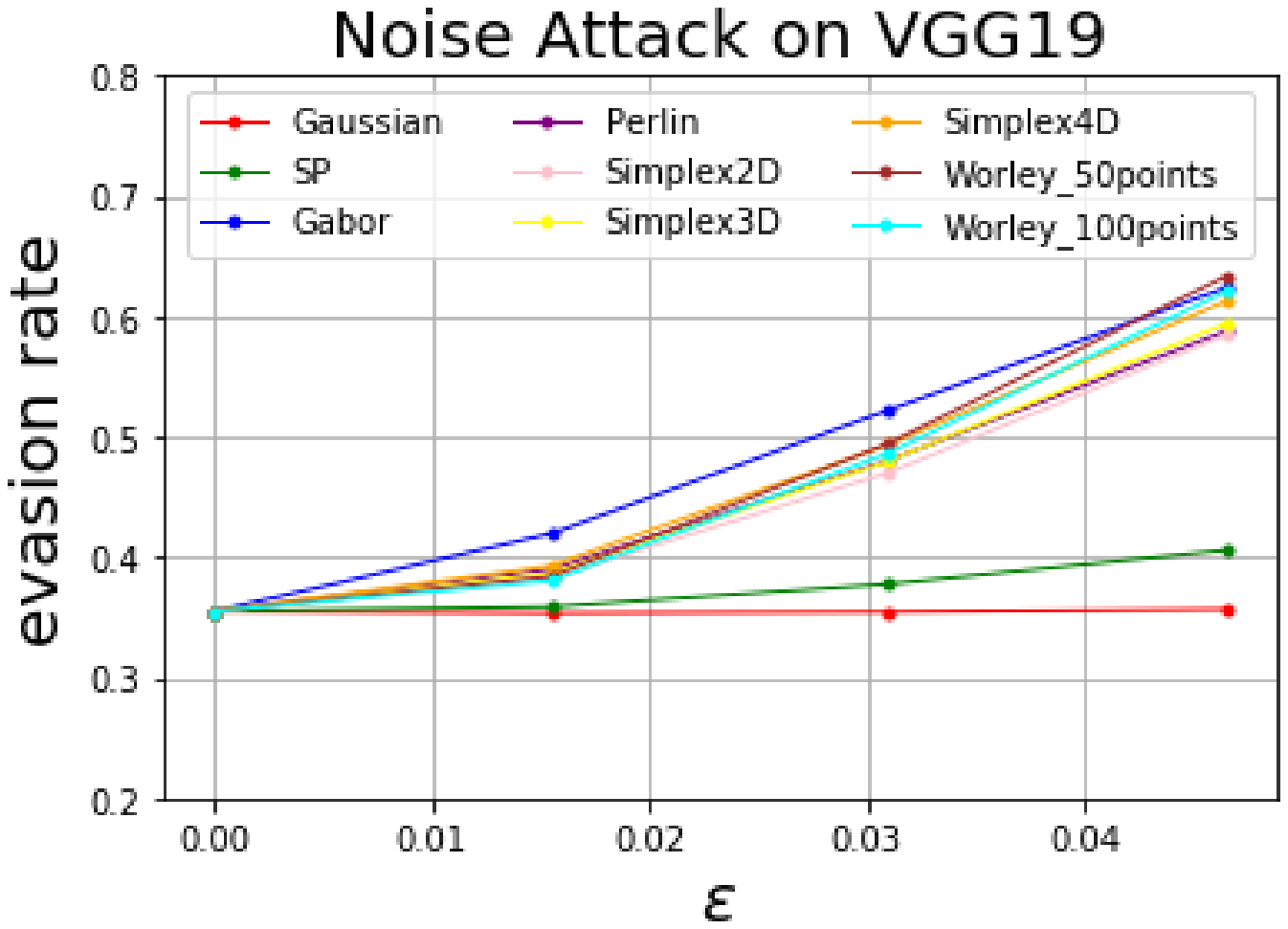}
}%
\\
\subfigure[]{
\centering
\includegraphics[width=0.4\hsize]{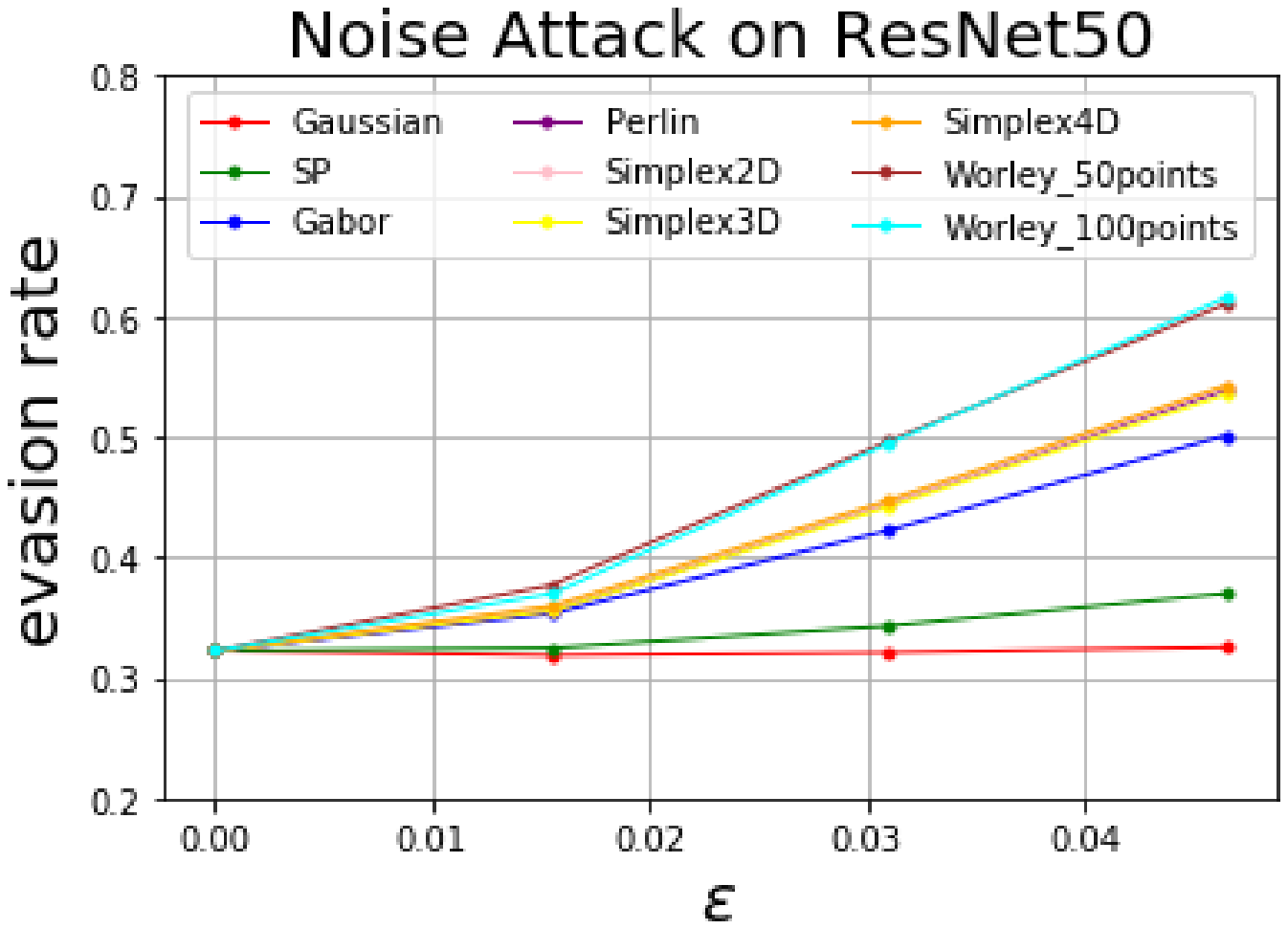}
}
\subfigure[]{
\centering
\includegraphics[width=0.4\hsize]{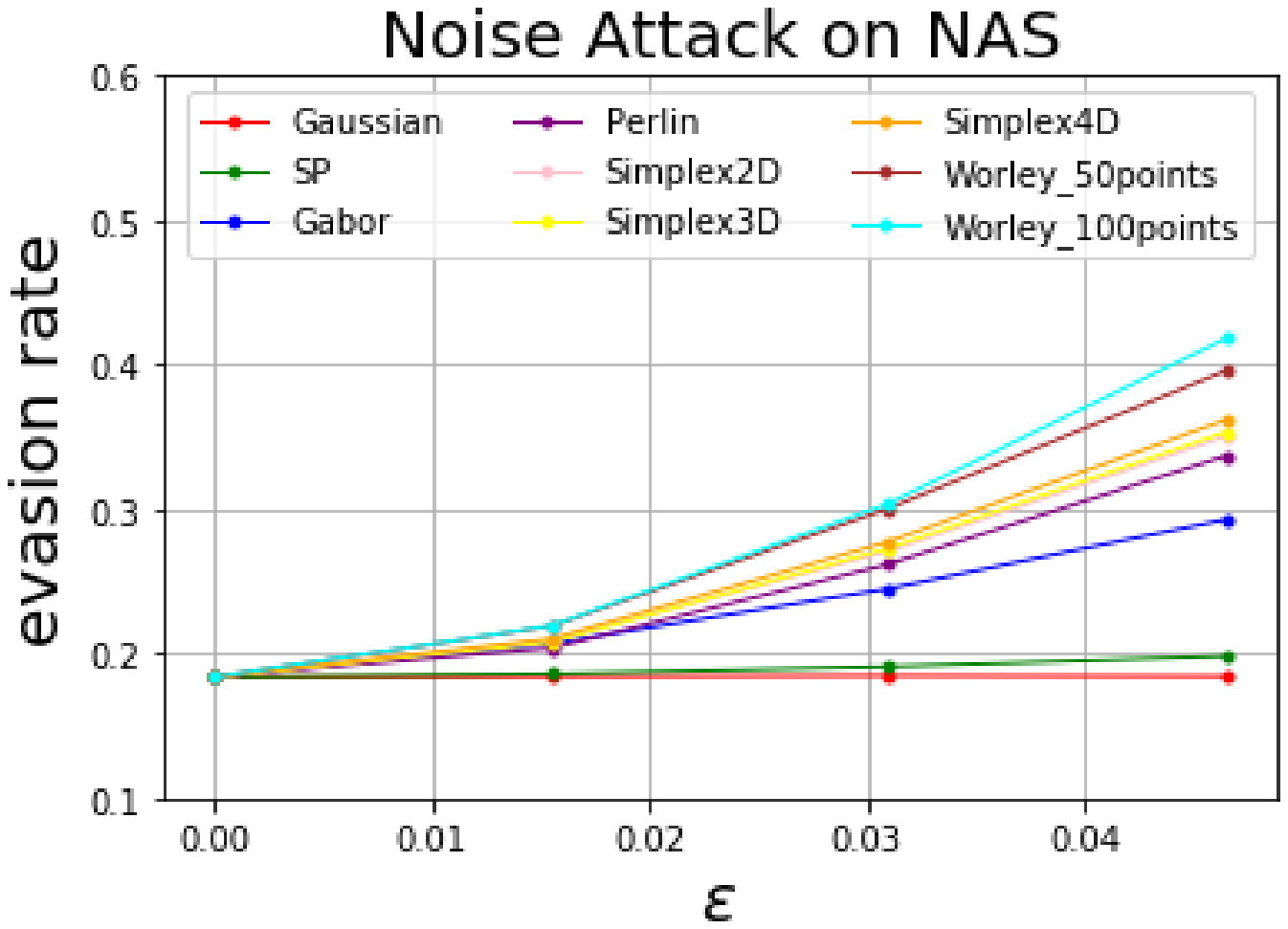}
}%
\caption{Experiment of procedural adversarial noise attack on ImageNet dataset.
Fig. (a) illustrates the experiment of procedural adversarial noise attack on InceptionV3~\citep{InceptionV3}.
Fig. (b) illustrates the experiment of procedural adversarial noise attack on VGG19~\citep{VGG}.
Fig. (c) illustrates the experiment of procedural adversarial noise attack on ResNet50~\citep{ResNet}.
Fig. (d) illustrates the experiment of procedural adversarial noise attack on NAS~\citep{NAS}.}
\label{fig:NoiseAttackImageNet}
\end{figure*}
\begin{figure*}[!htbp]
\centering
\subfigure[]{
\centering
\includegraphics[width=0.4\hsize]{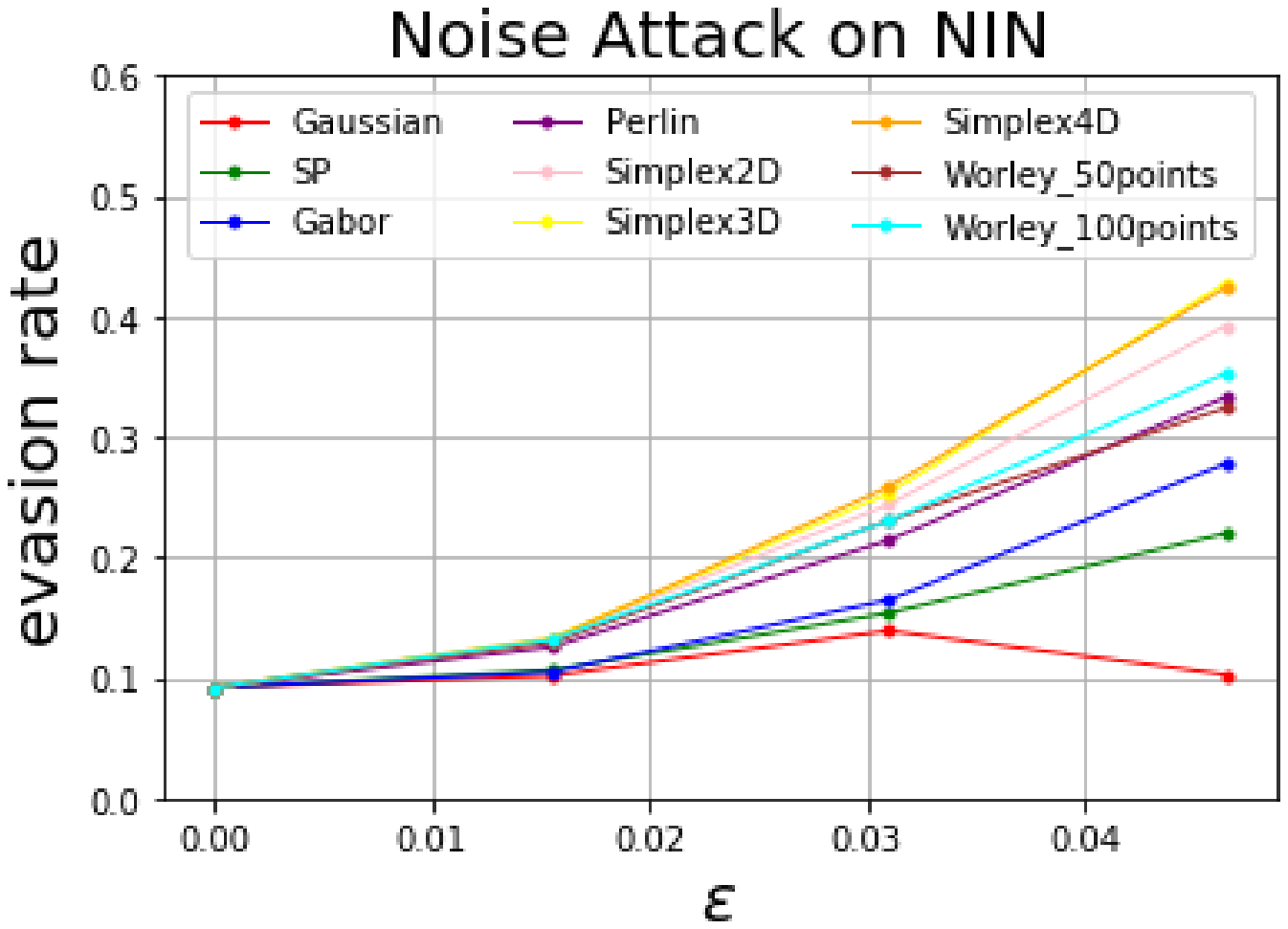}
}%
\subfigure[]{
\centering
\includegraphics[width=0.4\hsize]{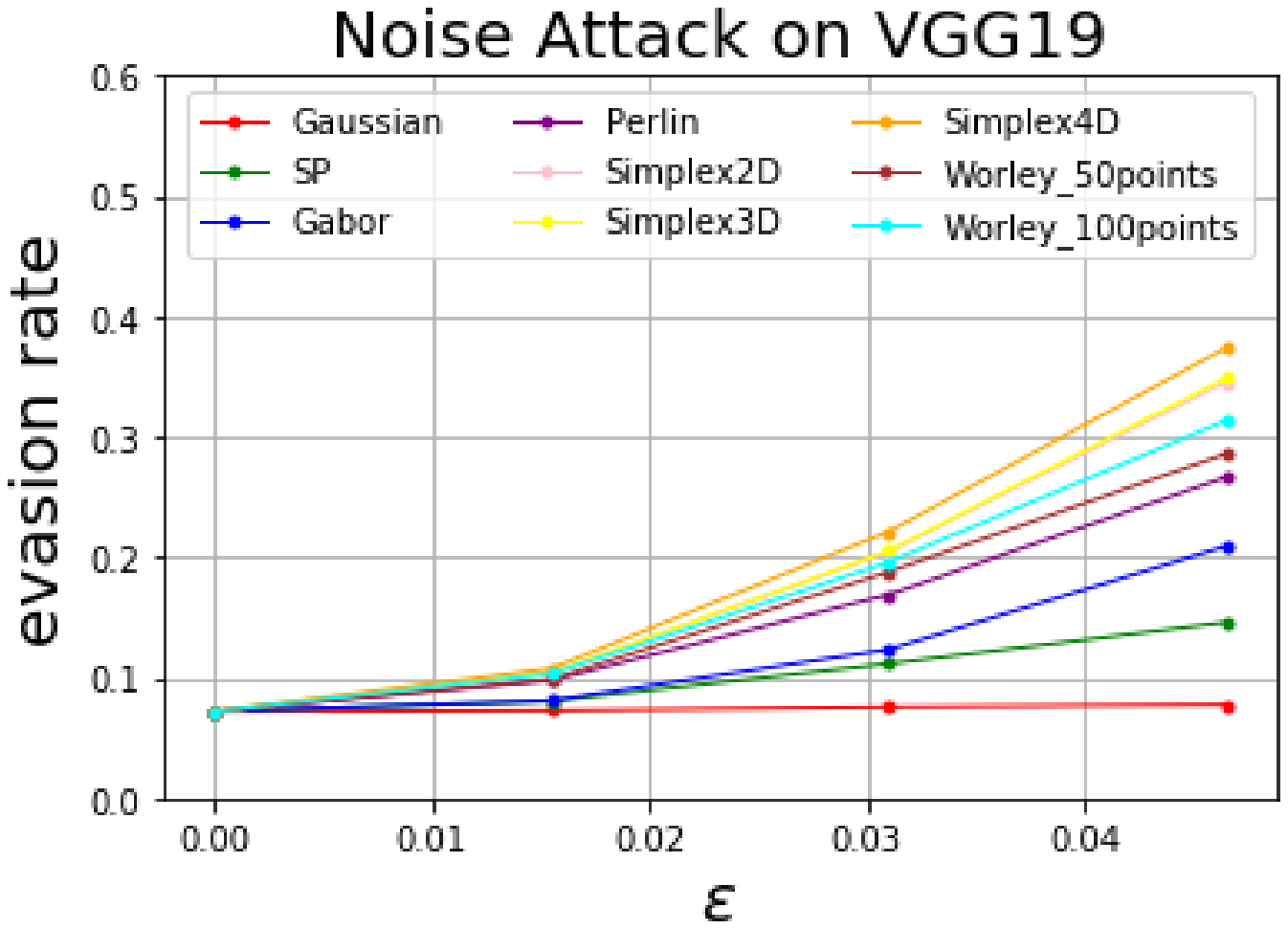}
}%
\\
\subfigure[]{
\centering
\includegraphics[width=0.4\hsize]{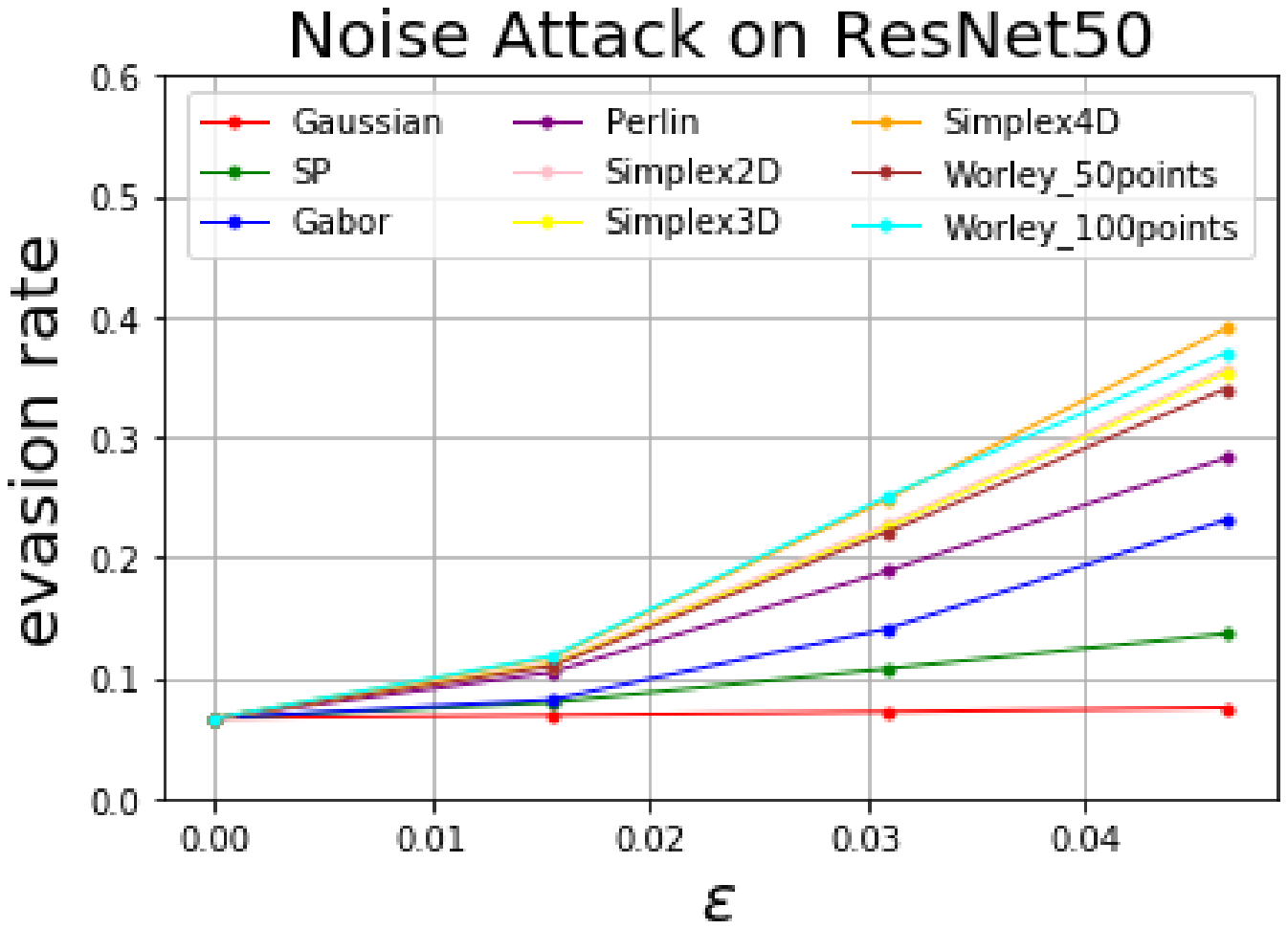}
}
\subfigure[]{
\centering
\includegraphics[width=0.4\hsize]{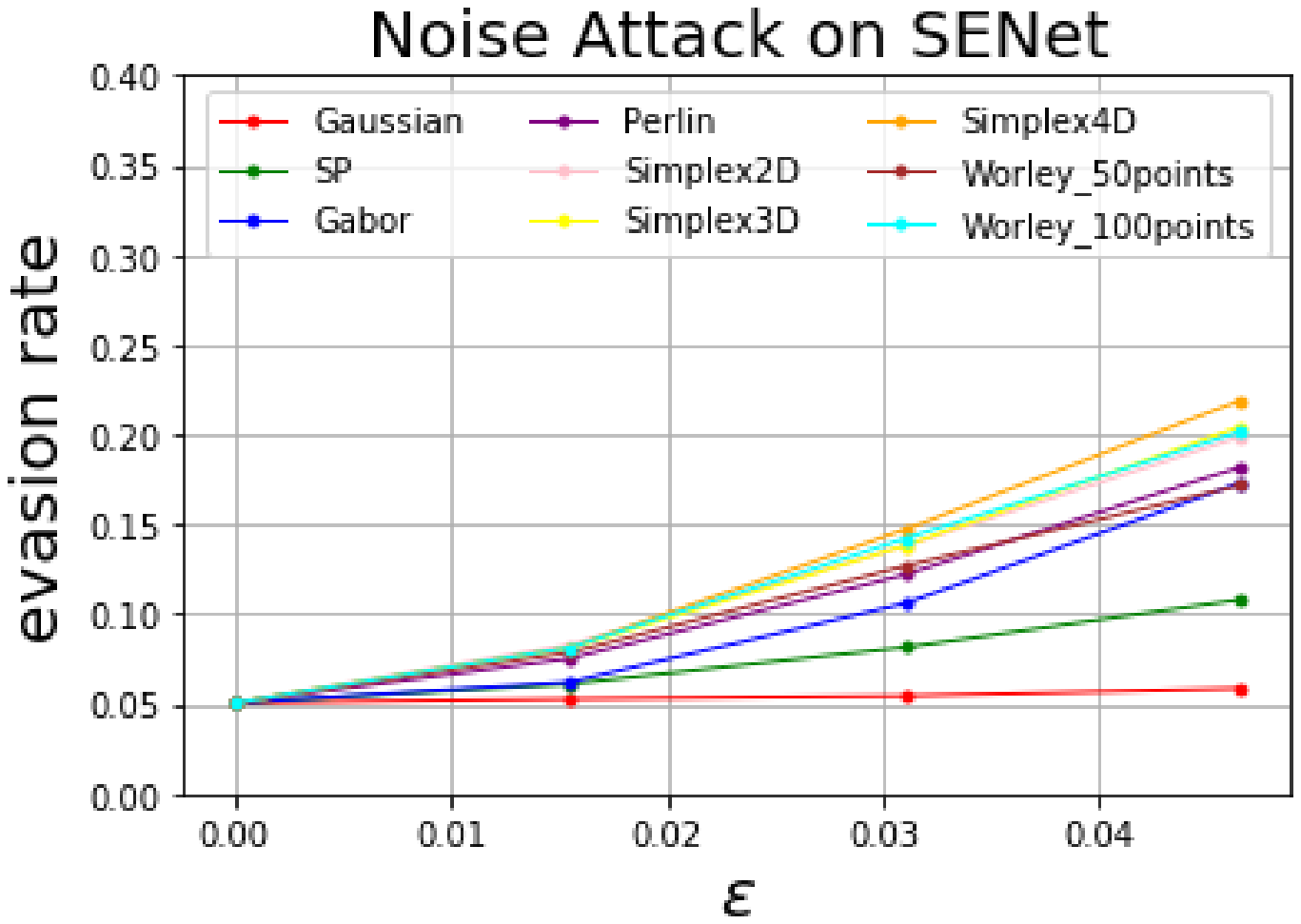}
}%
\caption{Experiment of procedural adversarial noise attack on the CIFAR-10 dataset.
Fig. (a) illustrates the experiment of procedural adversarial noise attack on NIN~\citep{NIN}.
Fig. (b) illustrates the experiment of procedural adversarial noise attack on VGG19~\citep{VGG}.
Fig. (c) illustrates the experiment of procedural adversarial noise attack on ResNet50~\citep{ResNet}.
Fig. (d) illustrates the experiment of procedural adversarial noise attack on SENet~\citep{SENet}.}
\label{fig:NoiseAttackCIFAR-10}
\end{figure*}
We do the procedural adversarial noise attack experiment on ImageNet with four models: InceptionV3~\citep{InceptionV3}, VGG19~\citep{VGG}, ResNet50~\citep{ResNet}, and neural architecture search model (NAS)~\citep{NAS}. They correspond to different convolutional neural networks: network-in-network structure, direct-connected structure, residual structure, and NAS model. Therefore, the cross-model attack performance of the procedural adversarial noises can be evaluated. All these models are pre-trained models inside Keras. We only check the top-1 prediction result.

Noise types on ImageNet are listed below. The perturbations budgets $\varepsilon$ of $\ell_{\infty}$-norm attack are set to $0.0155,\ 0.031,\ 0.0465$ for different norm attacks while $\varepsilon=0$ means natural testing without adversarial attack:
 \begin{itemize}
 \item \textbf{Gaussian noise (line mark is ``Gaussian")}: It is generated with the normal distribution whose mean value is 10 while the standard value is 50.

 \item \textbf{Salt-And-Pepper noise (line mark is ``SP")}: It is generated with Salt-And-Pepper noise on the probability of 0.1

 \item \textbf{Gabor noise~\citep{Gabor} (line mark is ``Gabor")}: It is generated with Gabor kernels whose kernel size is 23, its kernel variable $\sigma$, kernel orientation variable $\theta$, and bandwidth $\lambda$ are $8,\ 0.79,\ 8$.

 \item \textbf{Perlin noise~\citep{Perlin} (line mark is ``Perlin")}: The number of octaves is $\Omega \in[1,4]$, while period $T$ is 60, and frequency for  $\phi_{\text {sine }}$ function is 36.

 \item \textbf{Simplex noise generated in 2D dimensions (line mark is ``Simplex2D")}: It is iterated with the step of 40 to generate lattice gradient perturbations on 2D dimensions to produce Simplex noise.

 \item \textbf{Simplex noise generated in 3D dimensions (line mark is ``Simplex3D")}: It is iterated with the step of 40 to generate lattice gradient perturbations on 3D dimensions to produce Simplex noise.

 \item \textbf{ Simplex noise generate in 4D dimensions  (line mark is ``Simplex4D")}: It is iterated with the step of 40 to generate lattice gradient perturbations on 4D dimensions to produce Simplex noise.

 \item \textbf{Worley noise impacted on 50 points (line mark is ``Worley\_50points")}: It randomly clusters 50 points to generate Worley noise.

 \item \textbf{Worley noise impacted on 100 points  (line mark is ``Worley\_100points")}: It randomly clusters 100 points to generate Worley noise.
 \end{itemize}
The experiment result is illustrated in Fig. \ref{fig:NoiseAttackImageNet}. When the perturbation budget $\varepsilon$ of $\ell_{\infty}$-norm attack is 0.0465, the evasion rates for our proposed methods on Inception-V3 are 0.4935 (Simplex2D), 0.4895 (Simplex3D), 0.5065 (Simplex4D), 0.5929 (Worley\_50points), and 0.6336 (Worley\_100points). On VGG-19 with the same perturbation budget setting ($\varepsilon=0.0465$), the evasion rates are 0.5846 (Simplex2D), 0.5940 (Simplex3D), 0.6124 (Simplex4D), 0.6334 (Worley\_50points), and 0.6206 (Worley\_100points). On ResNet-50 ($\varepsilon=0.0465$), the result is 0.5414 (Simplex2D), 0.5352 (Simplex3D), 0.5427 (Simplex4D), 0.6102 (Worley\_50points), and 0.6155 (Worley\_100points). The NAS model has a best robustness performance under the $\ell_{\infty}$-norm noise attack with the perturbation budget $\varepsilon=0.0465$:  0.3505 (Simplex2D), 0.3505 (Simplex3D), 0.3617 (Simplex4D), 0.3957 (Worley\_50points), and 0.4181 (Worley\_100points).

On CIFAR-10, Network-in-network model (NIN)~\citep{NIN}, VGG19~\citep{VGG}, ResNet50~\citep{ResNet}, and SENet~\citep{SENet} are trained by ourselves.  These four models correspond to network-in-network structure, direct-connected structure, residual structure, and attention model. 
All the noise attack procedural design is similar to the experiment on ImageNet, except that the Simplex noise will be generated with iteration step 4. 
Also, the perturbations budgets $\varepsilon$ of $\ell_{\infty}$-norm attack are set to $0.0155,\ 0.031,\ 0.0465$ for different norm attacks while $\varepsilon=0$ means natural testing without adversarial attack. The experiment result is illustrated in Fig. \ref{fig:NoiseAttackCIFAR-10}. When the perturbation budget $\varepsilon$ of $\ell_{\infty}$-norm attack is 0.0465, the evasion rates for our proposed methods on NIN are 0.3456 (Simplex2D), 0.3486 (Simplex3D), 0.3738 (Simplex4D), 0.3242 (Worley\_50points), and 0.3528 (Worley\_100points). On VGG-19 with the same perturbation budget setting ($\varepsilon=0.0465$), the evasion rates are 0.3456 (Simplex2D), 0.3486 (Simplex3D), 0.3738 (Simplex4D), 0.2859 (Worley\_50points), and 0.3135 (Worley\_100points). On ResNet-50 ($\varepsilon=0.0465$), the result is 0.3564 (Simplex2D), 0.3528 (Simplex3D), 0.3900 (Simplex4D), 0.3401 (Worley\_50points), and 0.3698 (Worley\_100points). The SE-Net with the attention mechanism has a best robustness performance under the $\ell_{\infty}$-norm noise attack with the perturbation budget $\varepsilon=0.0465$:  0.1987 (Simplex2D), 0.2044 (Simplex3D), 0.2190 (Simplex4D), 0.1717 (Worley\_50points), and 0.2016 (Worley\_100points).

We can obtain some meaningful summaries of the attack experiment:

1) Our proposed procedural adversarial attack methods surpass state-of-the-art methods. Worley noise's evasion rate exceeds Simplex noise's evasion rate a little in the same condition on ImageNet, however, Simplex noise demonstrates a superior attack performance on CIFAR-10.

2) On ImageNet, as Fig. \ref{fig:NoiseAttackImageNet} demonstrates, NAS~\citep{NAS} is least sensitive to all types of adversarial attacks which testifies the value of research on neural architecture search and automated machine learning (AutoML). This can be verified in our experiment result.

3) On CIFAR-10, as Fig. \ref{fig:NoiseAttackCIFAR-10} shows, SENet~\citep{SENet} with channel attention is least sensitive to all the black-box noise attack. As we can see from the experiment result, the evasion rates on SENet do not surpass 25\%. It perhaps accords with some guess that the attention mechanism is beneficial to robustness. 
\subsection{Comparison Experiment of Black-Box Adversarial Attack}
 \begin{figure}[htbp]
\centering
\includegraphics[scale=0.5]{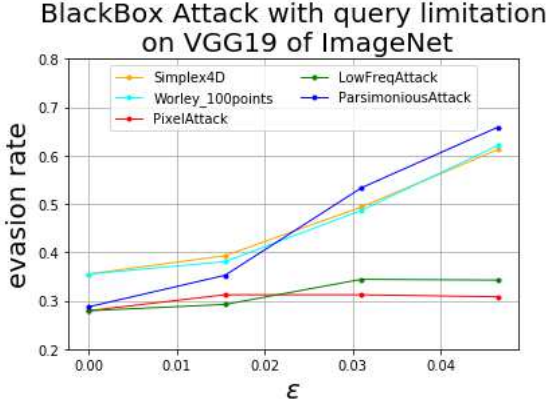}
\caption{Black-box adversarial attacks on VGG19~\citep{VGG} with query limitation on ImageNet.}
\label{fig:BBoxQueryAttackImageNet}
\end{figure}
In this subsection, we compare our methods with the query-based black-box attack methods~\citep{SimpleAttack, ParsimoniousAttack} in the query-limitation settings and transfer-based black-box attack methods~\citep{PGD, BIM, MIM} to show the superior performance of our proposed restricted black-box attack methods.

On ImageNet, VGG19~\citep{VGG} is the model to be attacked. Five listed methods are compared in the query-limited settings. 
 \begin{itemize}
  \item \textbf{ Simplex noise generate in 4D dimensions  (line mark is "Simplex4D")}: It is iterated with the step of 40 to generate lattice gradient perturbations on 4D dimensions to produce Simplex noise.
   \item \textbf{Worley noise impacted on 100 points  (line mark is "Worley\_100points")}: It randomly clusters 100 points to generate Worley noise.
   \item \textbf{Simple pixel attack (line mark is "PixelAttack")~\citep{SimpleAttack}}: The black-box attack is launched in the pixel level, while the query number is limited to 1000.
   \item \textbf{Simple low frequency attack (line mark is "LowFreqAttack")~\citep{SimpleAttack}}: The attack can be implemented in the frequency domain with DCT transform, while the query number is limited to 1000.
   \item \textbf{Parsimonious black-box attack via combinatorial optimization method (line mark is "ParsimoniousAttack")~\citep{ParsimoniousAttack}}: This black-box attack is realized by an efficient discrete surrogate to the combinatorial optimization problems, while the query number is limited to 500. 
 \end{itemize}
As Fig. \ref{fig:BBoxQueryAttackImageNet} illustrates, our proposed Simplex noise attack method ("Simplex4D") and Worley noise attack method ("Worley\_100points") outperform other state-of-the-art methods on the metrics of evasion rate if the query numbers of query-based black-box attack methods are limited to the specific scope.

In the transfer-setting experiment, we select 10000 samples from the ImageNet validation dataset. We generate the white-box adversarial examples via the method of PGD~\citep{PGD}, BIM~\citep{BIM}, MIM~\citep{MIM} on InceptionV3~\citep{InceptionV3} and attack the VGG19 model~\citep{VGG}. 
Due to the limitations of the computation power, we only test the scenario when  perturbation budget of $\ell_{\infty}$-norm attack $\varepsilon=0.0465$. TABLE \ref{tab:attackImageNetBboxTransfer} illustrates the result that our proposed Simplex noise attack method ("Simplex4D") and Worley attack method ("Worley\_100points") outperform the there compared methods regardless their high attack success rate (around 0.9) of InceptionV3 model~\citep{InceptionV3}. Simplex noise and Worley noise can be attributed to the universal adversarial perturbation that the attack can be transferred without the huge performance degradations between different model structures. However, the transfer-based adversarial attack methods based on the white-box attacks are affected by the performance decay.
\begin{table}[htbp]
\centering
\caption{COMPARISON OF BLACK-BOX ADVERSARIAL ATTACK IN THE TRANSFER SETTINGS ON IMAGENET.}
\label{tab:attackImageNetBboxTransfer}

\setlength{\tabcolsep}{0mm}{
\begin{tabular}{cc}
\toprule [2pt]
\textbf{Attack methods}  & \textbf{Evasion rate } \\
\hline
PGD & 0.2916  \\
\hline
BIM & 0.3943  \\
\hline
MIM  & 0.3998 \\
\hline
Simplex4D (our method) & 0.6124  \\
\hline
Worley\_100points (our method) &0.6206  \\

\bottomrule [2pt]

\end{tabular}}

\end{table}
\begin{figure}[htbp]

\centering
\includegraphics[scale=0.5]{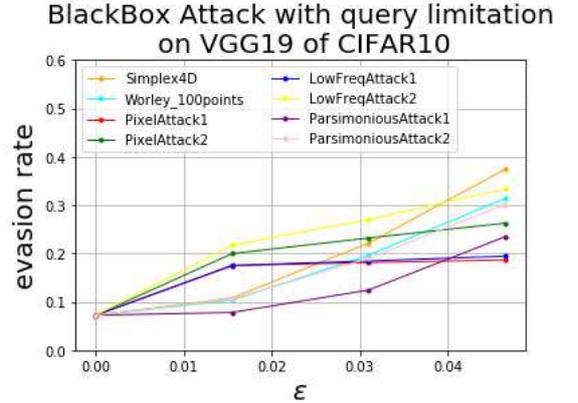}
\caption{Black-box adversarial attacks on VGG19~\citep{VGG} with query limitation on CIFAR-10. }
\label{fig:BBoxQueryAttackCIFAR}
\end{figure}
\begin{figure}[htbp]
\centering
\includegraphics[scale=0.5]{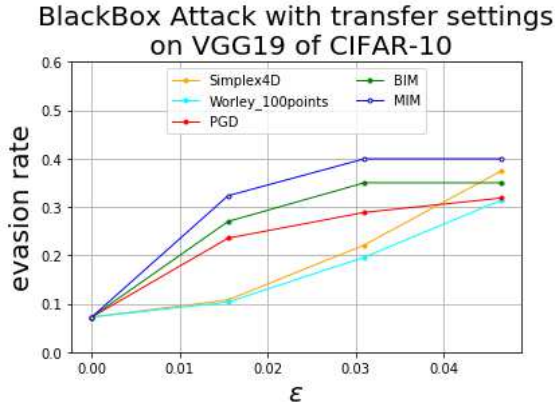}
\caption{Black-box adversarial attacks on VGG19~\citep{VGG} in transfer-settings. }
\label{fig:NewBBoxTransferAttackCIFAR}
\end{figure}

On CIFAR-10, we compare our proposed methods with the the state-of-the-art black-box attack methods~\citep{SimpleAttack, ParsimoniousAttack} in the query-limitation setting. The perturbation budget of $\ell_{\infty}$-norm attack $\varepsilon$ are set to $0.0155,\ 0.031,\ 0.0465$. The testing methods include:
 \begin{itemize}
  \item \textbf{ Simplex noise generate in 4D dimensions  (line mark is "Simplex4D", out method)}: It is iterated with the step of 4 to generate lattice gradient perturbations on 4D dimensions to produce Simplex noise.
   \item \textbf{Worley noise impacted on 100 points  (line mark is "Worley\_100points", our method)}: It is randomly clustered with 100 points to generate Worley noise.
   \item \textbf{Simple pixel attack (line mark is "PixelAttack1" and "PixelAttack2")~\citep{SimpleAttack}}: The black-box attack is launched in the pixel level,  the "PixelAttack1" method is with query-times of 100 while the "PixelAttack2" method is with the query-times of 500.
   \item \textbf{Simple low-frequency attack (line mark is "LowFreqAttack1" and "LowFreqAttack2")~\citep{SimpleAttack}}: The attack can be implemented in the frequency domain with DCT transform, the "LowFreqAttack1" method is with the query-times of 100 while the "LowFreqAttack2" method is with the query-time of 500.
   \item \textbf{Parsimonious black-box attack via combinatorial optimization method (line mark is "ParsimoniousAttack1" and "ParsimoniousAttack2")~\citep{ParsimoniousAttack}}: This black-box attack is realized by an efficient discrete surrogate to the combinatorial optimization problems, the "ParsimoniousAttack1" method is with the query-times of 100 while the "ParsimoniousAttack2" method is with the query-times of 200. 
 \end{itemize}
As Fig. \ref{fig:BBoxQueryAttackCIFAR} illustrates, our proposed Simplex noise attack method ("Simplex4D") and Worley noise attack method ("Worley\_100points") demonstrate superior performance when perturbation budget is not small ($\varepsilon \geq 0.031$).

In the transfer-setting experiment on CIFAR-10, we compare three black-box attack methods with the transfer from white-box adversarial examples: PGD~\citep{PGD}, BIM~\citep{BIM}, and MIM~\citep{MIM}. The adversarial examples are generated on the ResNet56~\citep{ResNet} with the attack training convergence, then the adversarial examples would be transferred to attack the VGG-19~\citep{VGG}. We find that the adversarial examples with 90 \% to 100 \% attack success rate on ResNet56 show inferior performance when attacking VGG19. The result is illustrated in Fig. \ref{fig:NewBBoxTransferAttackCIFAR}, which gives an empirical result that our proposed attack methods do not depend on model knowledge.

In summary, our proposed method surpasses the state-of-the-art methods on the metrics of evasion rate in the query-limitation setting and transfer setting.
\subsection{Comparison Experiment of Universal Adversarial Perturbations}
In this subsection, the metrics of evasion rate between our proposed methods and state-of-the-art universal perturbation generation methods~\citep{UAP, UAN, NAG, FastFeatureFool, GD_UAP} are compared.
On ImageNet, five different methods are compared on VGG-19~\citep{VGG} when perturbation budget of $\ell_{\infty}$-norm attack $\varepsilon$ is set to $0.04$:
 \begin{itemize}
  \item \textbf{ Simplex noise generate in 4D dimensions}: The setting is the same as the procedural adversarial noise attack experiment.
   \item \textbf{Worley noise impacted on 100 points}: The setting is the same as the procedural adversarial noise attack experiment.
   \item \textbf{Universal Adversarial Perturbation (UAP)~\citep{UAP}}: It is a vanilla universal adversarial generation method, which is data-driven.
   \item \textbf{Fast Feature Fool (FFF)~\citep{FastFeatureFool}}: It is a data-independent perturbation generation method with less calculation time, here would the adversarial examples generated on VGG16, VGG19, VGGF~\citep{VGG} and InceptionV1~\citep{InceptionV1}.
   \item \textbf{Generalizable data-independent Universal Adversarial Perturbation (GD-UAP)~\citep{GD_UAP}}: This method can be tested in three different settings: with full data, with range prior, and no data. The adversarial examples are generated on VGG-series models, InceptionV1, and ResNet152~\citep{ResNet}.
 \end{itemize}
 As illustrated in TABLE \ref{tab:attackImageNetUAP}, in specific settings, our data-independent universal perturbation generation methods with procedural adversarial noise functions surpass both the data-driven and data-independent UAP generation methods~\citep{UAP, FastFeatureFool, GD_UAP}.
 \begin{table*}[htbp]
\centering
\caption{COMPARISON OF UNIVERSAL ADVERSARIAL PERTURBATION METHODS ON VGG-19.}
\label{tab:attackImageNetUAP}

\setlength{\tabcolsep}{2mm}{
\begin{tabular}{cccc}
\toprule [2pt]
\textbf{Attack methods}  & \textbf{UAP-generation model} & \textbf{Data setting} & \textbf{Evasion rate } \\
\hline
 UAP &InceptionV1&With full data & 0.3992 \\
\hline
FFF-1& VGG-19 & No data & 0.5098  \\
\hline
FFF-2  & VGG-16 & No data & 0.5133\\
\hline
FFF-3  &VGGF&No data &  0.4971\\
\hline
FFF-4  &InceptionV1&No data & 0.5049  \\
\hline
GD-UAP & VGG-19 & No data & 0.5225 \\
\hline
GD-UAP & VGG-16 & No data & 0.5134\\
\hline
GD-UAP   &VGGF&No data &  0.5432\\
\hline
GD-UAP   &InceptionV1&No data & 0.4326  \\
\hline
GD-UAP & InceptionV1 & With full data & 0.5225 \\
\hline
GD-UAP  & InceptionV1 & With range prior & 0.5134\\
\hline
GD-UAP  &ResNet152 &No data &  0.4093\\
\hline
GD-UAP   &ResNet152 &With full data & 0.4955 \\
\hline
GD-UAP   &ResNet152 &With range prior & 0.4387  \\
\hline
Simplex4D (our method)  &- &No data & 0.5516 \\
\hline
Worley\_100points (our method)    &- &No data & 0.5598  \\
\bottomrule [2pt]

\end{tabular}}

\end{table*}

\begin{figure}[htbp]
\centering
\includegraphics[scale=0.5]{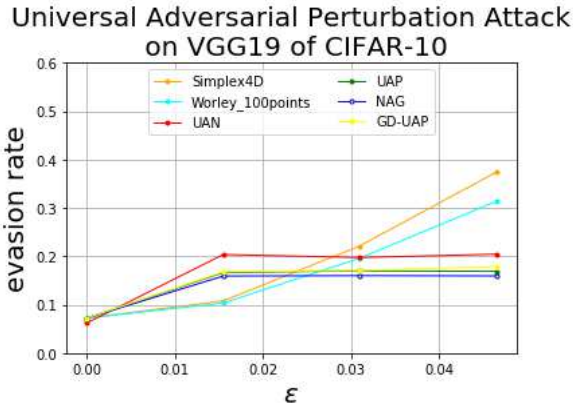}
\caption{Universal adversarial perturbation attacks on VGG19~\citep{VGG}  on CIFAR-10. }
\label{fig:uapCIFAR}
\end{figure}

On CIFAR-10, these listed methods are tested. It is worth mentioning that most proposed UAP methods do not provide an official baseline on CIFAR-10. This work reproduces the UAP algorithms on the CIFAR-10 dataset.
 \begin{itemize}
  \item \textbf{ Simplex noise generate in 4D dimensions}: The setting is the same as the procedural adversarial noise attack experiment.
   \item \textbf{Worley noise impacted on 100 points}: The setting is the same as the procedural adversarial noise attack experiment.
   \item \textbf{Universal adversarial perturbation (UAP)~\citep{UAP}}: It is a vanilla universal adversarial generation method, which is data-driven.
   \item \textbf{Generalizable data-independent Universal Adversarial Perturbation (GD-UAP)~\citep{GD_UAP}}: On CIFAR-10, only data-independent methods are tested.
    \item \textbf{Universal adversarial network (UAN)~\citep{UAN}}: The method is based on generative models to produce perturbations from a clean dataset.
   \item \textbf{Network for the adversarial generation (NAG)~\citep{NAG}}: The generative adversarial network is introduced to sample and produce perturbations.
   \end{itemize}
 As shown in Fig. \ref{fig:uapCIFAR}, our proposed method surpass the state-of-the-art UAP generation methods on CIFAR-10 when the perturbation budget satisfies $\varepsilon \geq 0.031$.
\subsection{Hyper-parameter Experiment}
In this sub-section, two hyper-parameters of our proposed methods would be checked on their performance metrics under different values. The iteration step is the hyper-parameter for the Simplex noise attack method, which is defined as $S$ in Algorithm \ref{alg:Simplex}. Perturbed point number is the hyper-parameter for Worley noise attack which is defined as $N$ in Algorithm \ref{alg:Worley}.

\begin{figure}[htbp]
\centering
\includegraphics[scale=0.5]{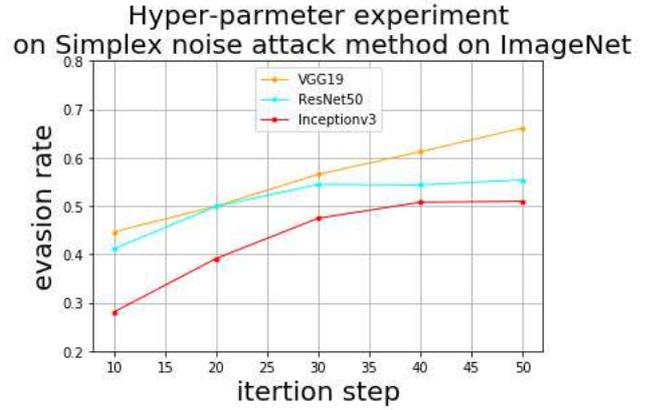}
\caption{Evasion rate of Simplex attack method with different iteration steps on ImageNet. }
\label{fig:HyperSimplexImageNet}
\end{figure}

\begin{figure}[htbp]
\centering
\includegraphics[scale=0.5]{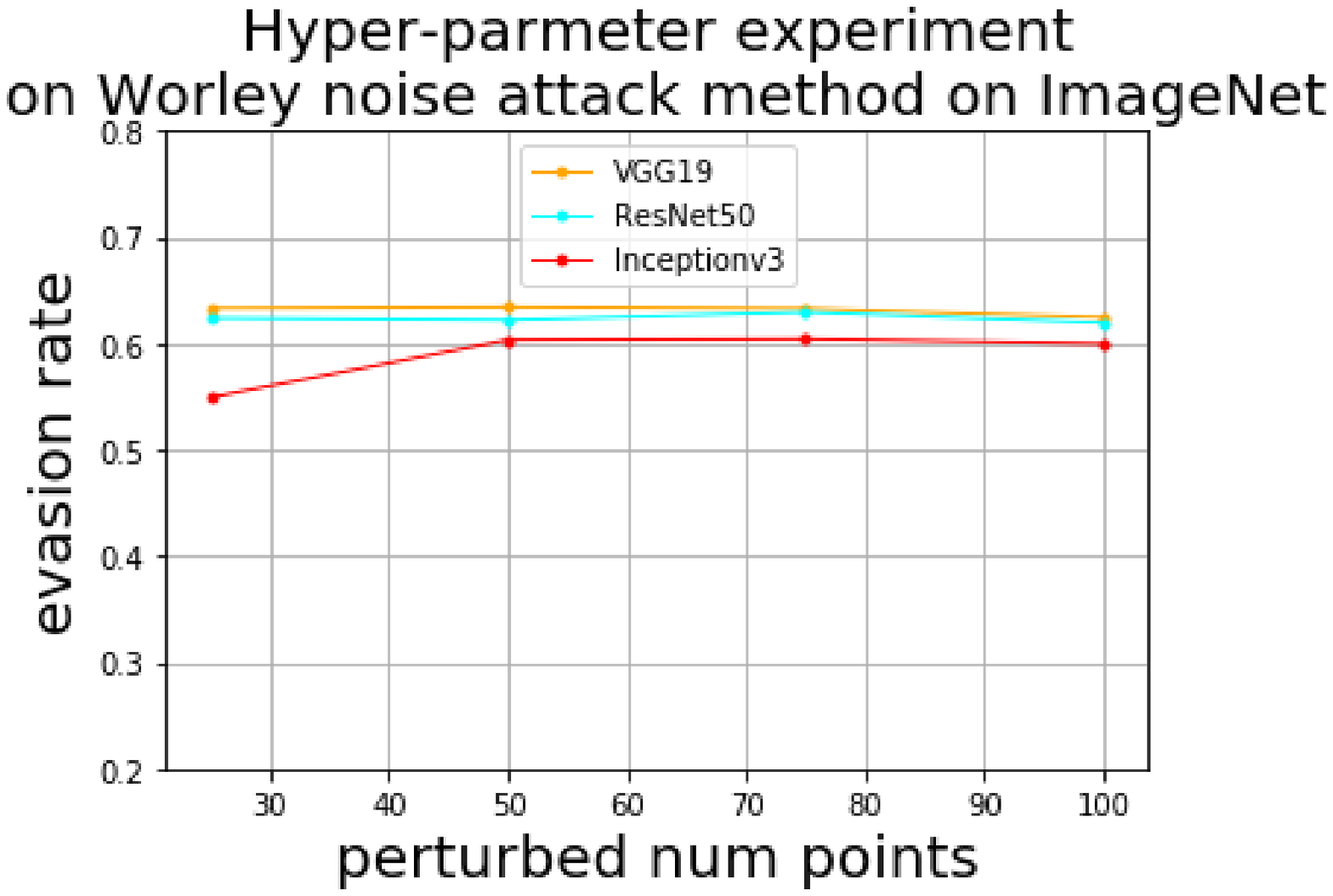}
\caption{Easion rate of Worley attack method with different perturbed point numbers on ImageNet.}
\label{fig:HyperWorleyImageNet}
\end{figure}

On ImageNet, we test the ``Simplex4D" scenario with different iteration steps while the perturbed point number in Worley noise attack method would be changed. The result of hyper-parameter experiment is illustrated in Fig. \ref{fig:HyperSimplexImageNet} and Fig. \ref{fig:HyperWorleyImageNet}.

\begin{figure}[htbp]
\centering
\includegraphics[scale=0.5]{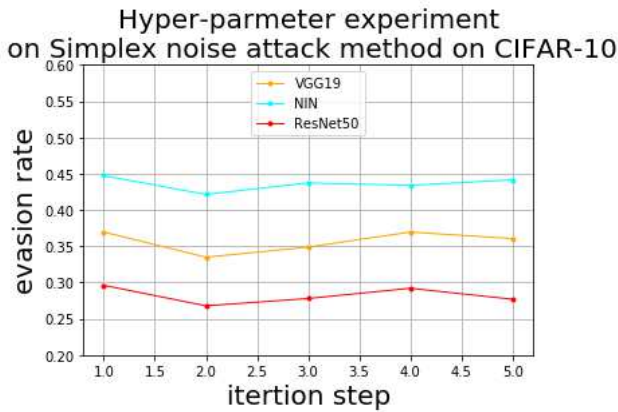}
\caption{Evasion rate of Simplex attack method with different iteration steps on CIFAR-10. }
\label{fig:HyperSimplexCIFAR}
\end{figure}

\begin{figure}[htbp]
\centering
\includegraphics[scale=0.5]{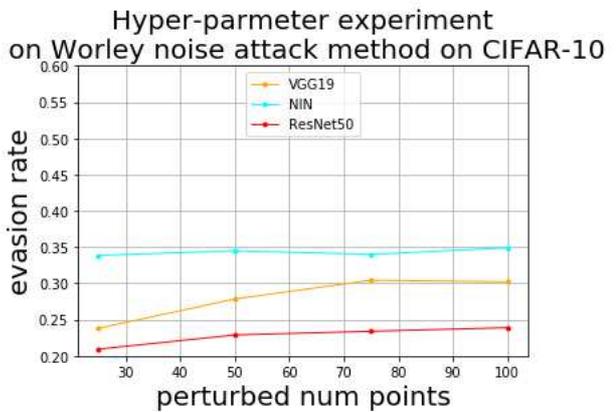}
\caption{Evasion rate of Worley attack method with different perturbed point numbers on CIFAR-10.}
\label{fig:HyperWorleyCIFAR}
\end{figure}

On CIFAR-10, we also test the ``Simplex4D" scenario with different iteration steps in a smaller search space while the perturbed point number in Worley noise attack method would vary. The result of hyper-parameter experiment is illustrated in Fig. \ref{fig:HyperSimplexCIFAR} and Fig. \ref{fig:HyperWorleyCIFAR}.

As can be seen from  Fig. \ref{fig:HyperSimplexImageNet}, Fig. \ref{fig:HyperWorleyImageNet}, Fig. \ref{fig:HyperSimplexCIFAR}, and Fig. \ref{fig:HyperWorleyCIFAR}, the iteration step size would matter in Simplex noise attack when the image size is large (e.g., ImageNet data). Otherwise, the settings of such parameters would not matter. 
\subsection{Experiment of Denoising-based Defense Methods}
\begin{figure*}[htbp]
\centering
\subfigure[]{
\centering
\includegraphics[width=0.35\hsize]{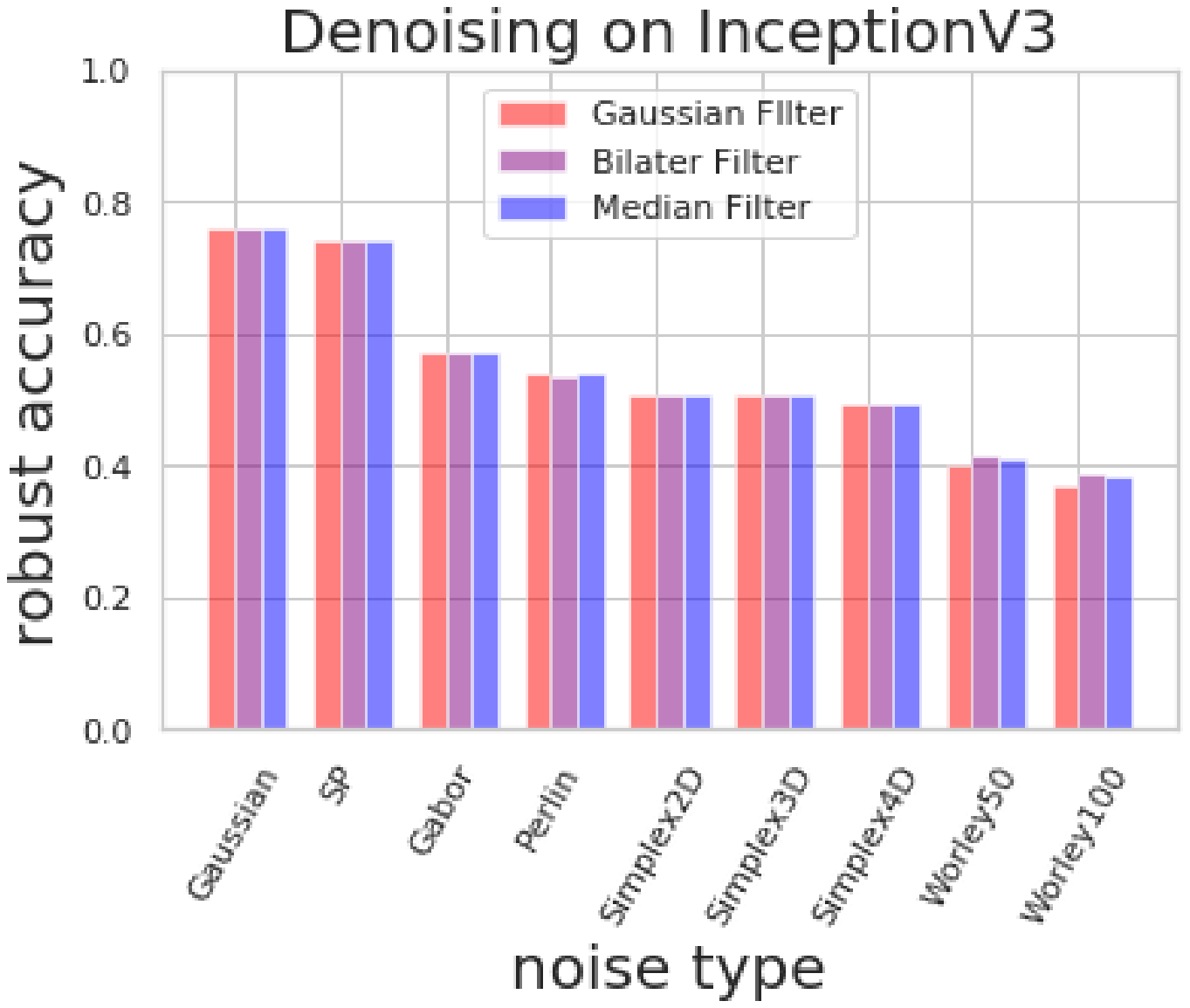}
}%
\subfigure[]{
\centering
\includegraphics[width=0.35\hsize]{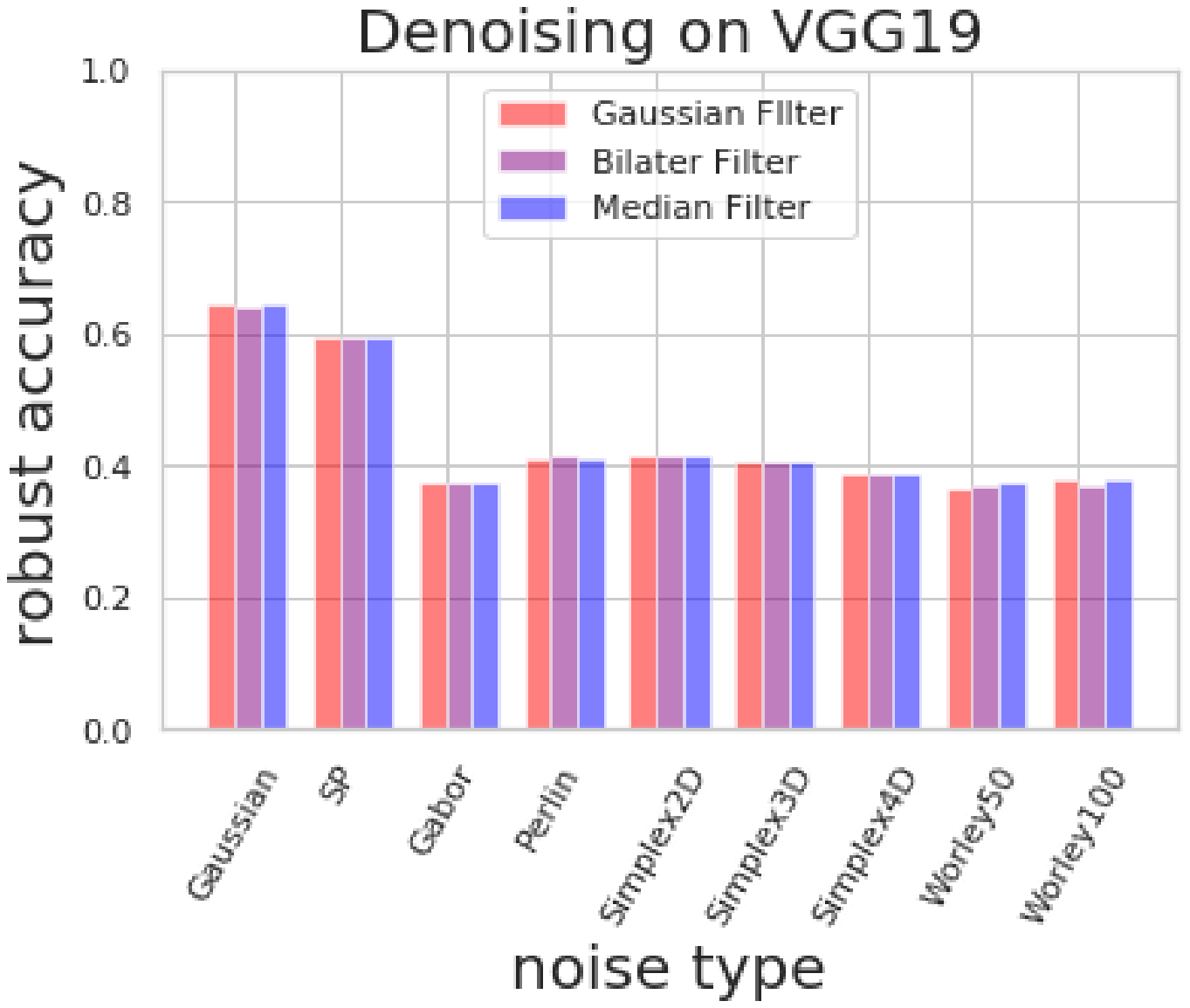}
}%
\\ 
\subfigure[]{
\centering
\includegraphics[width=0.35\hsize]{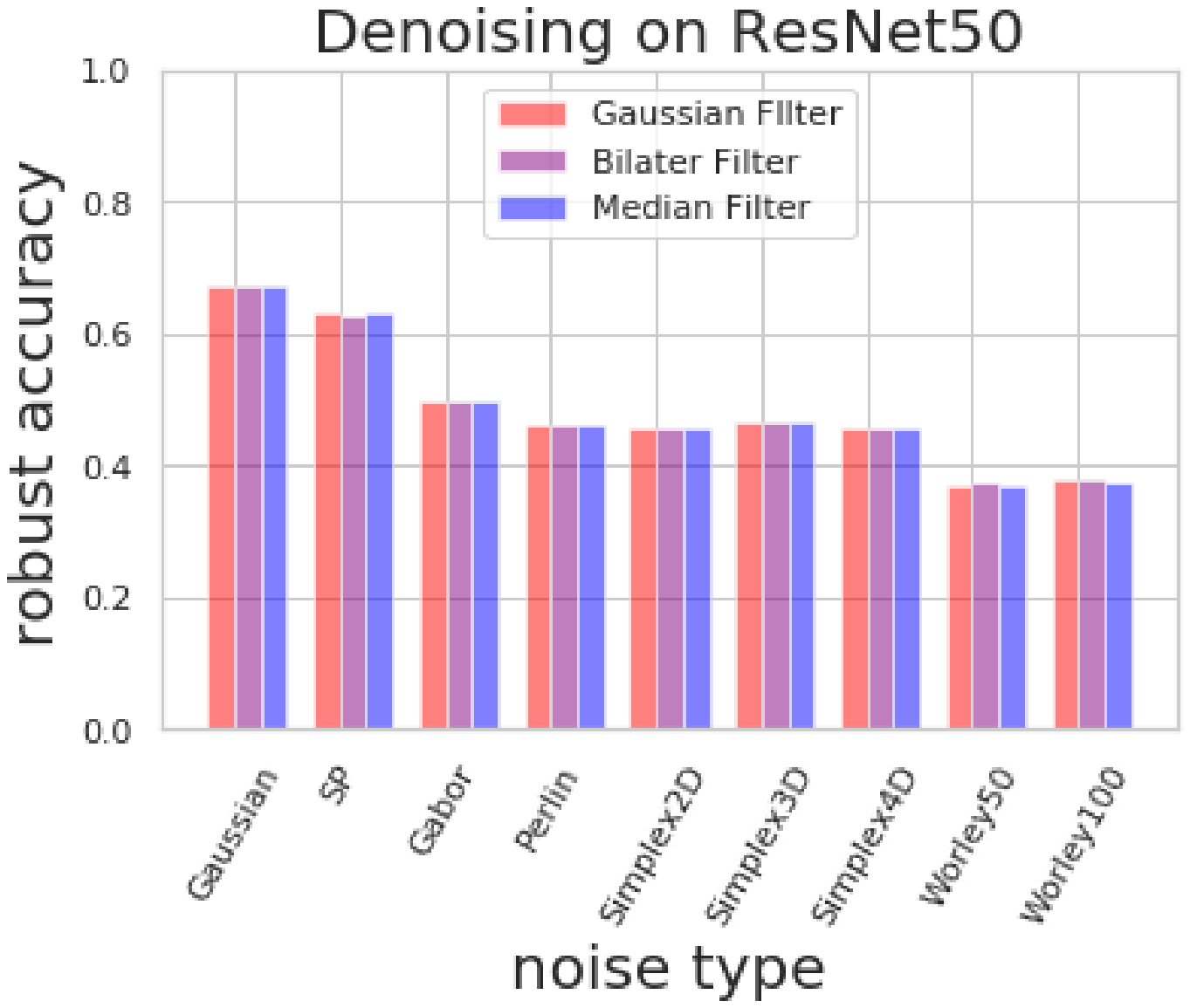}
}
\subfigure[]{
\centering
\includegraphics[width=0.35\hsize]{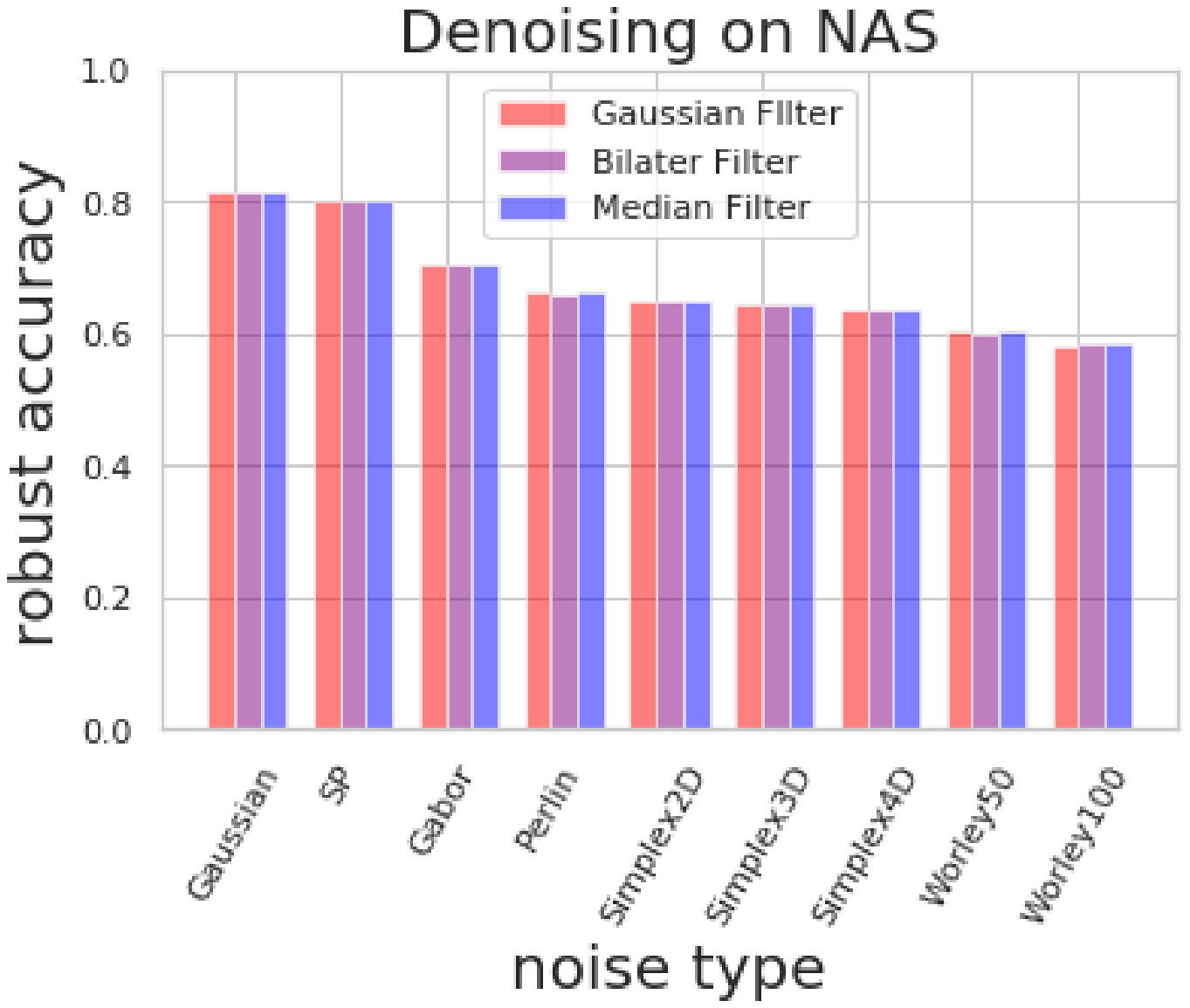}
}%
\caption{Experiment of denoising defense on the ImageNet dataset.
Fig. (a) illustrates the experiment of denoising methods combined with InceptionV3~\citep{InceptionV3} against adversarial noise attack.
Fig. (b) illustrates the experiment of denoising methods combined with VGG19~\citep{VGG} against adversarial noise attack.
Fig. (c) illustrates the experiment of denoising methods combined with ResNet50~\citep{ResNet}.
Fig. (d) illustrates the experiment of denoising methods combined with NAS~\citep{NAS}.}
\label{fig:DenoisingImageNet}
\end{figure*}
\begin{figure*}[htbp]
\centering
\subfigure[]{
\centering
\includegraphics[width=0.35\hsize]{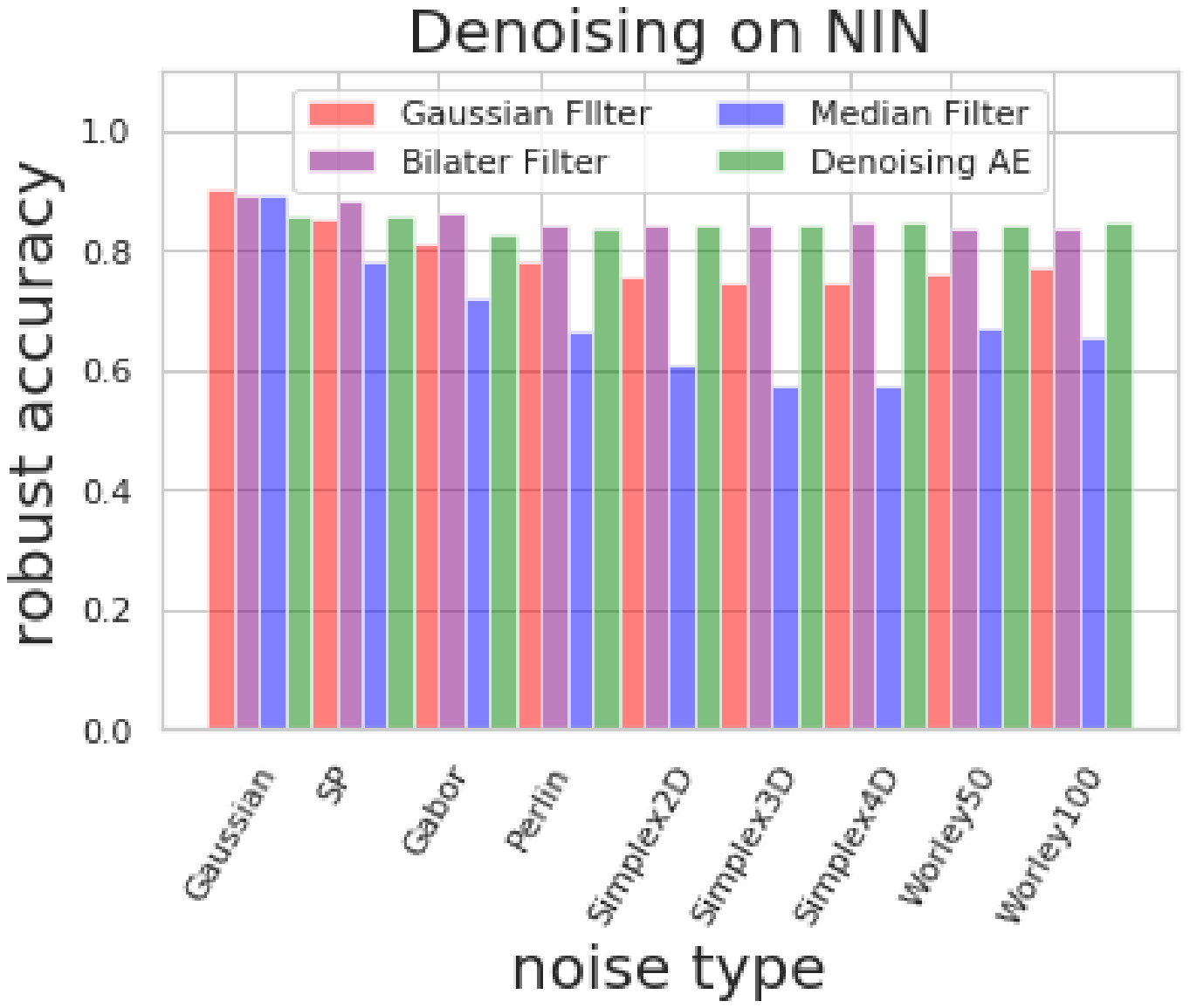}
}%
\subfigure[]{
\centering
\includegraphics[width=0.35\hsize]{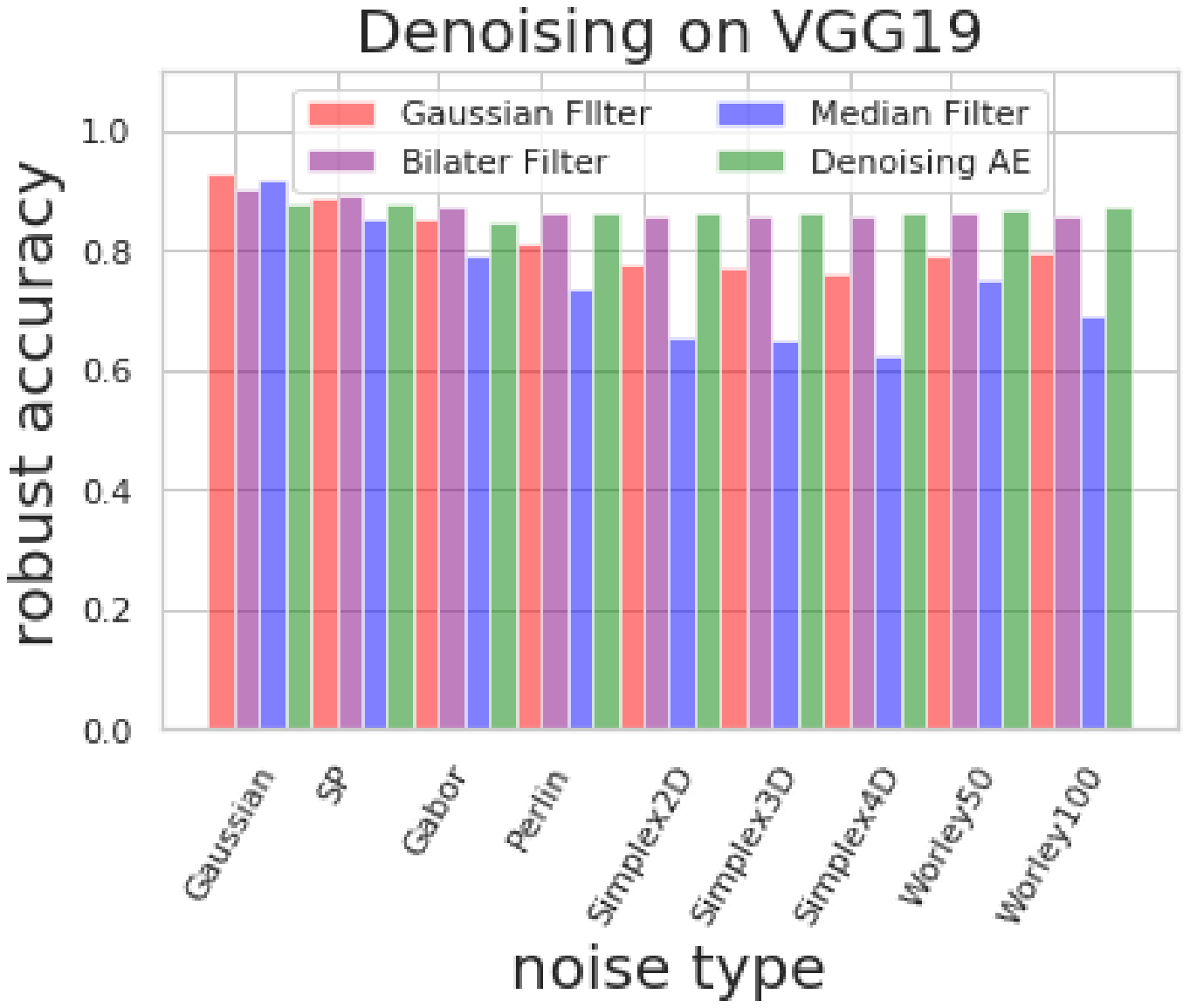}
}%
\\
\subfigure[]{
\centering
\includegraphics[width=0.35\hsize]{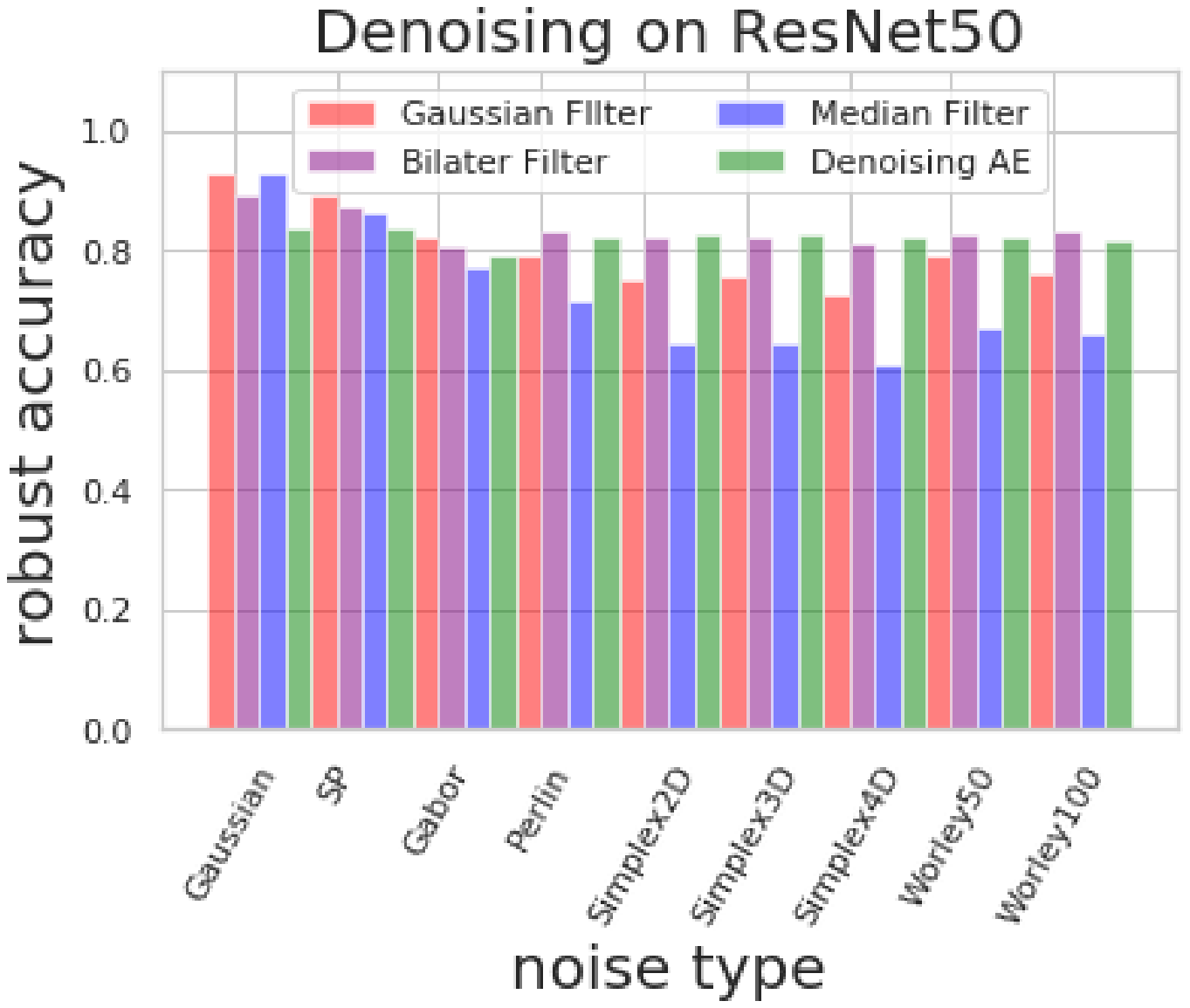}
}
\subfigure[]{
\centering
\includegraphics[width=0.35\hsize]{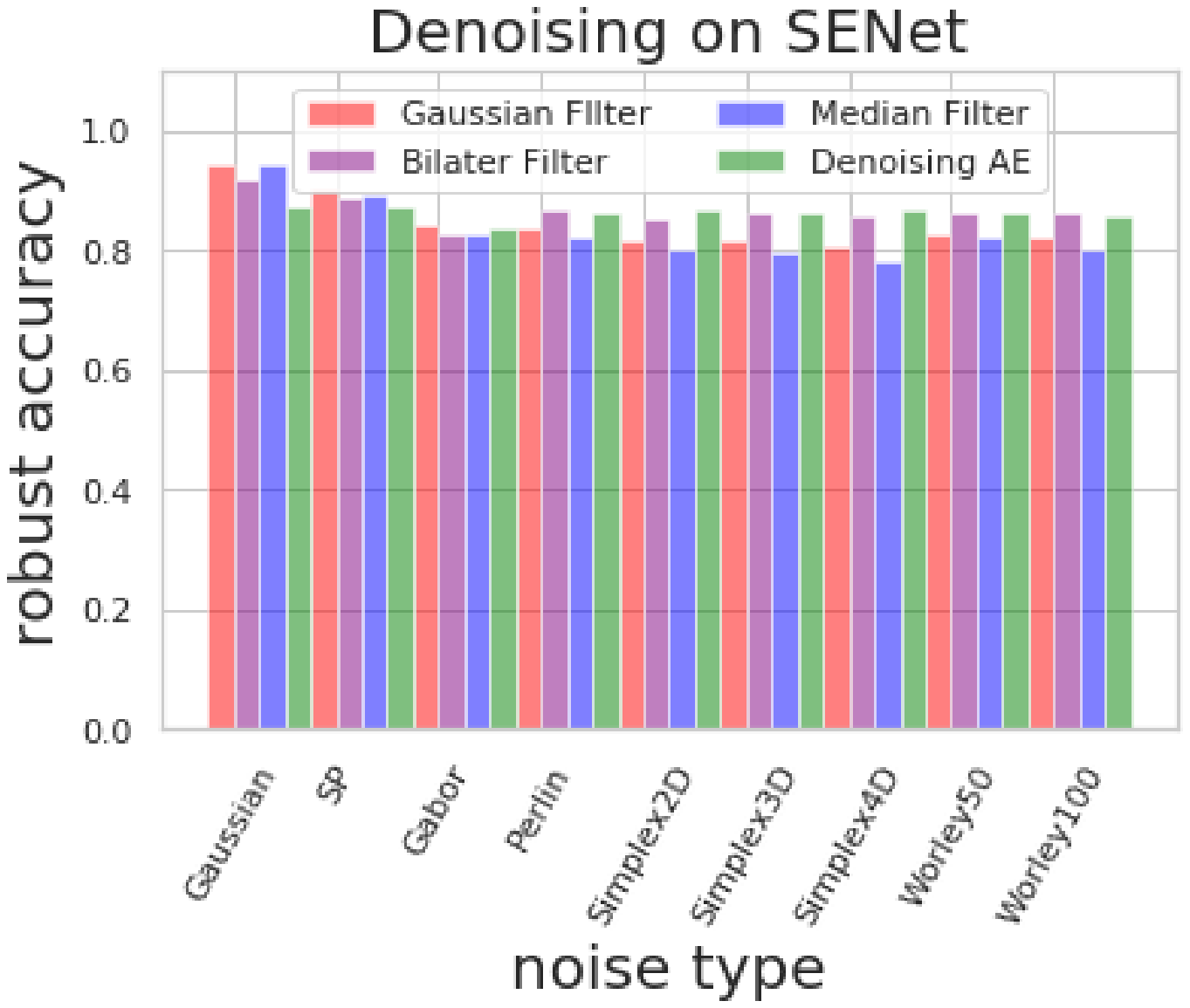}
}%
\caption{Experiment of denoising defense on CIFAR-10.
Fig. (a) illustrates the experiment of denoising methods combined with NIN~\citep{NIN} against adversarial noise attack.
Fig. (b) illustrates the experiment of denoising methods combined with VGG19~\citep{VGG} against adversarial noise attack.
Fig. (c) illustrates the experiment of denoising methods combined with ResNet50~\citep{ResNet}.
Fig. (d) illustrates the experiment of denoising methods combined with SENet~\citep{SENet}.}
\label{fig:DenoisingCIFAR10}
\end{figure*}
In the denoising experiment on ImageNet, we use the Gaussian filter, bilateral filter, and median filter for denoising defense. The attack type is set similar to the experiment of adversarial noise attack. We only test in the scenario when the perturbation budget of $\ell_{\infty}$-norm attack $\varepsilon$ is set to 0.0465.
The experiment result is showed in Fig. \ref{fig:DenoisingImageNet}, in which we can obtain the coarse conclusion that denoising methods have no effect on the ImageNet dataset. It perhaps accords the hypothesis~\citep{AdvInputDimension, Adversarial_Inevitable, AdvSpheres} that the neural network with large class numbers are inclined to be more vulnerable.

The result on CIFAR-10 with $\varepsilon=0.0465$ is completely different. As Fig. \ref{fig:DenoisingCIFAR10} illustrates, bilateral filtering and denoising autoencoder (denoising AE) are the two best methods that can guarantee robust accuracy. The F-Principle perhaps works when the image class number is not large because the bilateral-filter-based denoising method shows the superior performance on the defense under the procedural adversarial noise attack.

In summary, denoising methods are ineffective for complex image classes in high-dimensional spaces, however, when the image class number is not big, denoising methods with AE or bilateral filtering are helpful to improve the performance of robust accuracy. Taken VGG-19 as an example, when the perturbation budget of $\ell_{\infty}$-norm attack $\varepsilon$ is set to 0.0465, the robust accuracies with the bilateral filtering are: 0.8590 (Simplex2D), 0.8596 (Simplex3D), 0.858 (Simplex4D), 0.8599 (Worley\_50points), and 0.8595 (Worley\_100points). The robust accuracies with the denoising AE on VGG-19 are: 0.8646 (Simplex2D), 0.8616 (Simplex3D), 0.8635 (Simplex4D), 0.8688 (Worley\_50points), and 0.8732 (Worley\_100points). Compared with the AE-based denoising method, the bilateral-filter-based denoising method does not require additional training process. The experiment result verifies the F-Principle that the robustness of the neural networks is related to both the low-frequency elements and a little high-frequency elements.
\subsection{Experiment of Defense Methods in the RealSafe~\citep{RealSafe} Benchmark}
In this experiment, we keep the adversarial noise setting in the denoising experiment.

On ImageNet, the experiment is implemented on Inception v3 (Inc-v3)~\citep{InceptionV3}, ensemble adversarial training (Ens-AT)~\citep{EnsembleAdversarialTraining}, adversarial logit pairing (ALP)~\citep{ALP}, feature denoising (FD)~\citep{DenoisingAdversarialTraining}, JPEG compression (JPEG)~\citep{JpegDefense}, bit-depth reduction (Bit-Red)~\citep{BitReduction}, random resizing and padding (RandomResizingPadding, RP)~\citep{RP}, and RandMix~\citep{RandMix}. The experiment result is illustrated in Table \ref{table:RealSafeImageNet}. As we can see, the model performance of Ens-AT~\citep{EnsembleAdversarialTraining} does not degrade too much under most adversarial noise attacks except Worley noise, the performance of denoising training model~\citep{DenoisingAdversarialTraining} is stable with the cost of test accuracy.
\begin{table*}[htbp]
\caption{DEFENSE METHODS OF REALSAFE~\citep{RealSafe} AGAINST PROCEDURAL ADVERSARIAL NOISE ATTACKS ON IMAGENET}
\label{table:RealSafeImageNet}
\resizebox{\textwidth}{!}{
\centering
\begin{tabular}{lccccccccccc}
\toprule
     & Gaussian  &   SP & Gabor &  Perlin &   Simplex2D &  Simplex3D  &  Simplex4D  &  Worley50 & Worley100 \\  
 \midrule
Inc-v3  &0.7804 & 0.7722 & 0.5785 & 0.5343 & 0.7290 & 0.7270 & 0.7330 & 0.3293 & 0.3228 \\
Ens-AT & \textbf{0.7413} & \textbf{0.7234} & \textbf{0.6239} & \textbf{0.6473} & \textbf{0.7060} & \textbf{0.7050} & \textbf{0.7150} & \textbf{0.5270} & \textbf{0.5593} \\
ALP  & 0.4844 & 0.4800 & 0.4598 & 0.4621 & 0.4620 & 0.4540 & 0.4580 & 0.3855 & 0.4616   \\
FD & 0.6422 & 0.6413 & 0.6300 & 0.6342 & 0.6320 & 0.6330 & 0.6330 & 0.6326 & 0.6328 \\
JPEG & 0.7705 & 0.7591 & 0.5578 & 0.5546 & 0.7140 & 0.7030 & 0.7040 & 0.3311 & 0.3442 \\
Bit-Red & 0.6576 & 0.6523 & 0.5340 & 0.5541 & 0.6920 & 0.6770 & 0.6760 & 0.4398 & 0.4533 \\
RP & 0.7512 & 0.7280 & 0.5402 & 0.5198 & 0.6938 & 0.6894 & 0.6887 & 0.3052 & 0.3191  \\
RandMix & 0.5330 & 0.5192 & 0.3432 & 0.4125 & 0.4997 & 0.4928 & 0.4934 & 0.3259 & 0.3576 \\
         \bottomrule
\end{tabular}}
\end{table*}

On CIFAR-10, the defense models include: ResNet-56 (Res-56)~\citep{ResNet}, PGD-based adversarial training (PGD-AT)~\citep{PGD}, DeepDefense~\citep{DeepDefense}, TRADES~\citep{TRADES}, convex outer polytope (ConvexDefense)~\citep{ConvexDefense}, JPEG compression (JPEG)~\citep{JpegDefense}, random self-ensemble (RSE)~\citep{RSE}, and adaptive diversity promoting (ADP)~\citep{ADP}.  The result is illustrated in Table \ref{table:RealSafeCIFAR10}, it seems that PGD-AT~\citep{PGD}, TRADES~\citep{TRADES}, and RSE~\citep{RSE} are robust against these attacks which means that adversarial training and ensemble are two effective methods that help improve the robustness of the neural network models.
\begin{table*}[htbp]
\caption{DEFENSE METHODS OF REALSAFE~\citep{RealSafe}  AGAINST PROCEDURAL ADVERSARIAL NOISE ATTACKS ON CIFAR-10}
\label{table:RealSafeCIFAR10}
\resizebox{\textwidth}{!}{
\centering
\begin{tabular}{lccccccccccc}
\toprule
     & Gaussian  &   SP & Gabor &  Perlin &   Simplex2D &  Simplex3D  &  Simplex4D  &  Worley50 & Worley100\\  
 \midrule
Res-56 &0.9177 & 0.8104 & 07620 & 0.6329 & 0.5054& 0.4787 & 0.4801 & 0.5604 & 0.516 \\
PGD-AT & \textbf{0.8716} & \textbf{0.8652} & \textbf{0.8519} & \textbf{0.8551} & \textbf{0.8547} & \textbf{0.8569} & \textbf{0.8622} & \textbf{0.8590} & \textbf{0.8597}\\
DeepDefense  & 0.7797 & 0.6457 & 0.6140 & 0.5283 & 0.4074 & 0.3076 & 0.4033 & 0.5445 & 0.4922  \\
TRADES& \textbf{0.8505} & \textbf{0.8431}& \textbf{0.8305} & \textbf{0.8299} & \textbf{0.8293} & \textbf{0.8301} & \textbf{0.8293} & \textbf{0.8281} & \textbf{0.8294} \\
Convex & 0.6579 & 0.6596 & 0.6524 & 0.6582 & 0.6591 & 0.6606 & 0.6590 & 0.6571 & 0.6590 \\
JPEG & 0.8859 & 0.7330 & 0.6693 & 0.5671 & 0.4359 & 0.4159 & 0.4139 & 0.5182 & 0.4371 \\
RSE& \textbf{0.8579} & \textbf{0.8575} & \textbf{0.8379} & \textbf{0.8535} & \textbf{0.8538} & \textbf{0.8551} & \textbf{0.8561} & \textbf{0.8554} & \textbf{0.8578} \\
ADP & 0.9325 & 0.8870 & 0.8262 & 0.7471 & 0.6306& 0.6131 & 0.6075 & 0.7404 & 0.6982 \\
         \bottomrule
\end{tabular}}
\end{table*}

We also test the defense methods on our proposed method and other universal adversarial perturbation generation methods: Simplex4D,  Worley100 (with 100 perturbed points), UAP~\citep{UAP}, FFF~\citep{FastFeatureFool}, and GD-UAP~\citep{GD_UAP} in three settings (data, free, range prior). The UAP-based adversarial examples are generated on InceptionV1~\citep{InceptionV1}, while FFF-based adversarial examples and GD-UAP examples are generated on VGG-19~\citep{VGG}. As TABLE \ref{table:RealSafeUAPImageNet} illustrates,  JPEG defense and ensemble adversarial training are two effective methods for defense on universal adversarial perturbations.
\begin{table*}[!htbp]
\caption{DEFENSE METHODS OF REALSAFE~\citep{RealSafe} AGAINST UNIVERSAL ADVERSARIAL PERTURBATIONS ON IMAGENET}
\label{table:RealSafeUAPImageNet}
\resizebox{\textwidth}{!}{
\centering
\begin{tabular}{lccccccc}
\toprule
     & UAP  &   FFF & GD-UAP (data) &  GD-UAP (free) &  GD-UAP (range prior) &  \  Simplex4D  &   Worley100 \\  
 \midrule
Inc-v3  &0.633&	0.569&	0.405&	0.564&	0.446&	0.733&	0.3228 \\
Ens-AT& \textbf{0.622} &	\textbf{0.585} &	\textbf{0.494}	& \textbf{0.572} & \textbf{0.523}	 & \textbf{0.71}	& \textbf{0.5593} \\
ALP & 0.339 &	0.282 &	0.292 &	0.26 &	0.293 &	0.458 &	0.4616   \\
FD~  & 0.479 &	0.442 &	0.436 &	0.444 &	0.429 &	0.633 &	0.6328 \\
JPEG&\textbf{0.6449}	& \textbf{0.584} &	\textbf{0.411} &	\textbf{0.576} &	\textbf{0.464} &	\textbf{0.704}  & \textbf{0.3442} \\
Bit-Red &0.537	& 0.488 &	0.407 &	0.473 &	0.444&	0.676 &	0.4533 \\
RP & 0.588 & 0.5279	 & 0.3958	& 0.5158 &	0.4375 &	0.6887 &	0.3191  \\
RandMix &0.4073 &	0.373 &	0.3297 &	0.3706 &	0.3435 &	0.4934 &	0.3576 \\
         \bottomrule
\end{tabular}}
\end{table*}

In summary, adversarial training methods~\citep{PGD, TRADES, EnsembleAdversarialTraining} are effective methods to defend the procedural adversarial noise attacks and universal adversarial perturbations.
\subsection{Discussion}
\begin{itemize}
\item \textbf{Why are the proposed two attacks effective for fooling neural networks?} In real world, human eyes would be interfered by the shadings. The procedural adversarial noise attack based on the computer graphics rendering technologies can generate shadings with the inspiration of the interference mechanism in human perception systems. It leads to the deception against the neural networks. Currently, there are no effective methods to separate and remove shadows~\citep{ShadowDetectionSurvey}. Moreover,  adding universal adversarial perturbations based on procedural noise functions is to augment the high-frequency elements which can cause uncertainty in the pixel domain. Last but not least, current computer vision algorithms are based on the mechanism of pixel recognition instead of global semantic understanding which leads to the vulnerabilities under the attacks. 

\item \textbf{The cost of the procedural adversarial noise attack.} Procedural adversarial noise attack is $l_{\infty}$-norm attack which aims to limit or minimize the amount that any pixel is perturbed to achieve an adversary’s goal. In some scenarios, using $l_{\infty}$-norm attack would affect the quality of the image, although it does not change the meaning of the image. In the paper, our proposed Simplex noise attack method and Worley noise attack method do not depend on the model's prior knowledge and achieve a considerable result with the metrics of evasion rate. The computation cost of Simplex noise is $O(n^{2})$ with $n$ dimensions compared to the $O(n \cdot 2^{n})$ of Perlin noise. However, to generate such procedural adversarial noises, the iteration on the whole image space is needed.
    
\item \textbf{What are potentially effective defense technologies?} Frequency Principle (F-Principle) can interpret the robustness of neural network models. As the result of the denoising defense experiments on CIFAR-10 illustrates, both low-frequency elements related to image features and high-frequency elements connected with robustness are important. The performance of the bilateral filtering denoising method based on bandpass filter surpasses the performance of the denoising methods based on Gaussian filter and median filter. Moreover, adversarial robustness can be realized via adversarial training. Madry et al.~\citep{PGD} pointed out that there exists a ``gradient penalty” phenomenon in adversarial training. Adding perturbations is the operation of gradient ascent which penalizes the descending gradient for optimization not to be too large. This ``gradient penalty” mechanism guarantees the robustness in deep learning. The empirical study on ImageNet validates the analysis further that the InceptionV3 model which is trained with augmented perturbations from diverse CNNs (Ens-AT)~\citep{EnsembleAdversarialTraining} is robust to the noises except Worley noise. It can also be seen as the evidence supporting the effectiveness of adversarial training.    
\end{itemize}

\section{Conclusion}
The research on universal adversarial perturbations (UAPs) is explorable. Procedural adversarial noise is one type of UAPs. In this paper, we propose two procedural adversarial noise attack methods to craft image-agnostic universal adversarial perturbations (UAPs): Simplex noise and Worley noise. Our proposed attack methods surpass the state-of-the-art procedural adversarial noise methods on ImageNet and CIFAR-10. An empirical study is made to compare our methods with other black-box adversarial attack methods and universal adversarial perturbation attack methods. The effectiveness of the adversarial noise attack method lies in the shading generated by the rendering technologies which disturbs the classification abilities of neural networks. Discomfort with the shading does not only exist in the machine vision systems but also in the human perception system. It raises a security challenge of the current deep-learning-based visual system. Moreover, an empirical study of the defense methods on the procedural adversarial noises is illustrated. The results of our defense experiments validate some theoretical analysis of robustness in deep learning. Several findings can be highlighted: 1) In the denoising-based defense experiment on CIFAR-10, the methods satisfied with the Frequency Principle (F-Principle) boost the robustness under the adversarial attack; 2) In the defense experiment of RealSafe benchmark, the adversarial training methods with ``gradient penalty" mechanism provides a robustness guarantee under the procedural adversarial noise attack. Our work provides a little inspiration for the research on universal adversarial perturbations (UAP). This may boost the research to improve the robustness of neural networks. 	

\begin{acknowledgements}
This work was supported by the National Natural Science Foundation of China under Grant No. 61701348. The authors would like to thank TUEV SUED for the kind and generous support.
\end{acknowledgements}

%
%

\bibliographystyle{spbasic}      


\bibliography{ijcv_resubmit_procedural_noise}
%
%

\end{sloppypar}
\end{document}